\documentclass[runningheads]{llncs}
\usepackage{graphicx}
\usepackage{comment}
\usepackage{amsmath,amssymb} 
\usepackage{color}
\usepackage{xpatch}

\usepackage{packages}

\begin{document}
\pagestyle{headings}
\mainmatter

\title{\TITLE} 


\titlerunning{\SHORTTITLE}
\newcommand*\samethanks[1][\value{footnote}]{\footnotemark[#1]}

\author{Kangning Liu\inst{1,2}\thanks{Equal contributions} \and Shuhang Gu\inst{2}\samethanks  \and Andr{\'e}s Romero\inst{2}  \and Radu Timofte\inst{2}
}

\authorrunning{Liu et al.}
%
\institute{Center for Data Science, New York University, USA \\
\and
Computer Vision Lab, ETH Z{\"u}rich,   Switzerland
}
\maketitle


\begin{abstract}
Existing unsupervised video-to-video translation methods fail to produce translated videos which are frame-wise realistic, semantic information preserving and video-level consistent. In this work, we propose UVIT, a novel unsupervised  video-to-video translation model.
Our model decomposes the style and the content, uses the specialized encoder-decoder structure and propagates the inter-frame information through bidirectional recurrent neural network (RNN) units.
The style-content decomposition mechanism enables us to achieve style consistent video translation results as well as provides us with a good interface for modality flexible translation.
In addition, by changing the input frames and style codes incorporated in our translation, we propose a video interpolation loss, which captures temporal information within the sequence to train our building blocks in a self-supervised manner. 
Our model can produce photo-realistic, spatio-temporal consistent translated videos in a multimodal way.
Subjective and objective experimental results validate the superiority of our model over existing methods. More details can be found on our project website: \url{https://uvit.netlify.com/}.

\end{abstract}

\section{Introduction}
\label{sec:introduction}

Recent image-to-image translation (I2I) methods have achieved astonishing results by employing Generative Adversarial Networks (GANs)~\cite{goodfellow2014generative}. 

While there is an explosion of papers on I2I, its video counterpart is much less explored. 
Nevertheless, the ability to synthesize dynamic visual representations is important to a wide range of tasks such as video colorization~\cite{zhang2019deep}, medical imaging~\cite{nie2016estimating}, model-based reinforcement learning~\cite{arulkumaran2017brief,ha2018world}, computer graphics rendering~\cite{kajiya1986rendering}, etc.

Compared with the I2I task, the video-to-video translation (V2V) is more challenging. Besides the frame-wise realistic and semantic preserving requirements, which are also required in the I2I task, V2V methods additionally need to consider the temporal consistency for generating sequence-wise realistic videos.
Consequently, directly applying I2I methods on each frame of the video is not an optimal solution because I2I cross-domain mapping does not hold temporal consistency within the sequence.

Recent methods~\cite{bansal2018recycle,bashkirova2018unsupervised,chen2019mocycle} have included different constraints to model the temporal consistency based on the Cycle-GAN approach~\cite{zhu2017unpaired} for unpaired datasets. 
They either use a 3D spatio-temporal translator~\cite{bashkirova2018unsupervised} or add a temporal loss on traditional image-level translator \cite{bansal2018recycle,chen2019mocycle}. However, 3DCycleGAN~\cite{bashkirova2018unsupervised} heavily sacrifices image-level quality, and RecycleGAN~\cite{bansal2018recycle} suffers from style shift and inter-frame random noise. 
\begin{figure}[t]
\vspace{-0.3cm}
    \centering
    \includegraphics[scale=1]{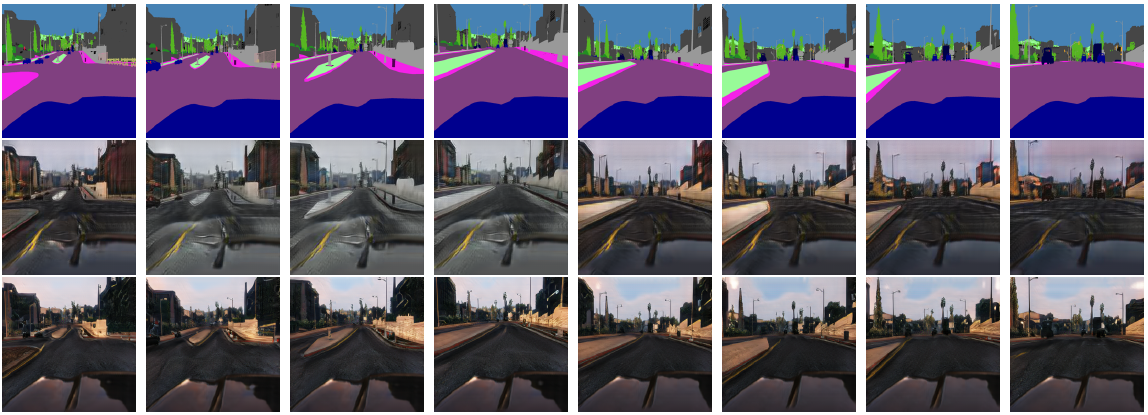}
    \vspace{-0.3cm}
    \caption{\textbf{A consistent video should be 1) style inconsistent 2) content consistent} First row: label inputs; Second row: ReCycleGAN\cite{bansal2018recycle} outputs; Third row: \SHORTTITLE{} (ours) outputs. To overcome the style shift (\emph{e.g. }sunset frame gradually changes to rain frame), we utilize style-conditioned translation. To reduce artifacts across frames, our translator incorporate multi-frame information. We use systematic sampling to get the results from a 64-frame sequence. The full video is provided in \AppendixName{}
    }
    \label{fig:compareHR}
    \vspace{-0.6cm}
\end{figure}

In this paper we propose \TITLE{} (\SHORTTITLE{}), a novel framework for video-to-video cross-domain mapping. To this end, a temporally consistent video sequence translation should simultaneously guarantee: \textit{(1)} Style consistency, and \textit{(2)} Content consistency, see~\fref{fig:compareHR} for a visual example. Style consistency requires the whole video sequence to have the same style, thus ensuring the video frames to be overall realistic. Meanwhile, content consistency refers to the appearance continuity of contents in adjacent video frames, which ensures the video sequence to be dynamically vivid.

\begin{figure}[t]
\begin{center}
\includegraphics[width=0.5\linewidth,trim={8 0 20 0},clip]{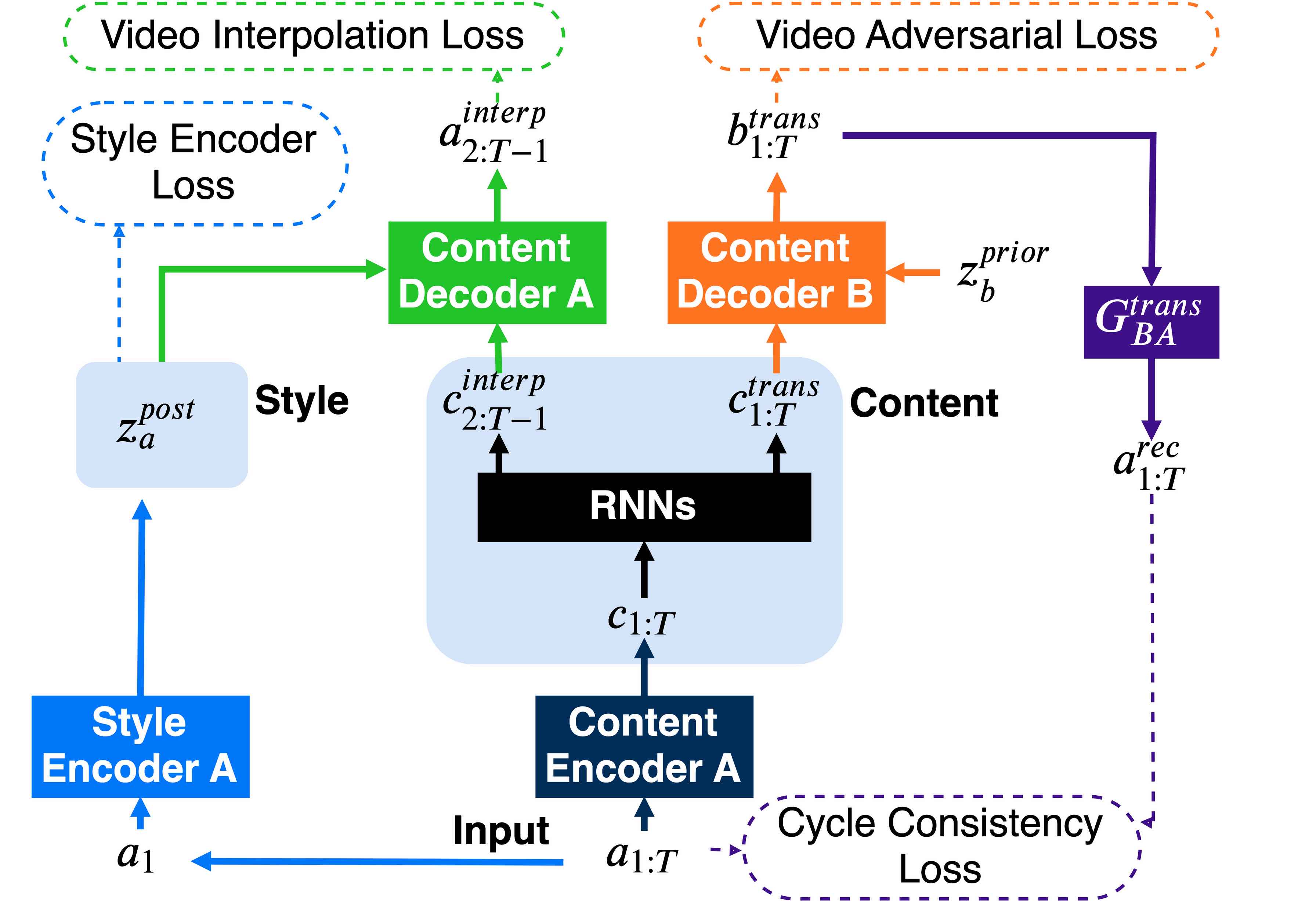}
\end{center}
\vspace{-7mm}
\caption{Overview of our proposed \SHORTTITLE{}~model: given an input video sequence, we first decompose it to the content by a Content Encoder and the style by a Style Encoder. Then the content is processed by special RNN units, namely TrajGRUs \cite{shi2017deep} in order to get the content used for translation and interpolation in a recurrent manner. Finally, the translation content and the interpolation content are decoded to the translated video and the interpolated video together with the style latent variable. We also show the video adversarial loss ({\color{orange} orange}), the cycle consistency loss ({\color{violet} violet}), the video interpolation loss ({\color{green!60!black} green}) and the style encoder loss ({\color{blue} blue})
}
\vspace{-0.3cm}
\label{fig:overview}
\end{figure}

\begin{figure}[ht]
\vspace{-0.3cm}
\centering
  \includegraphics[width=0.6\linewidth]{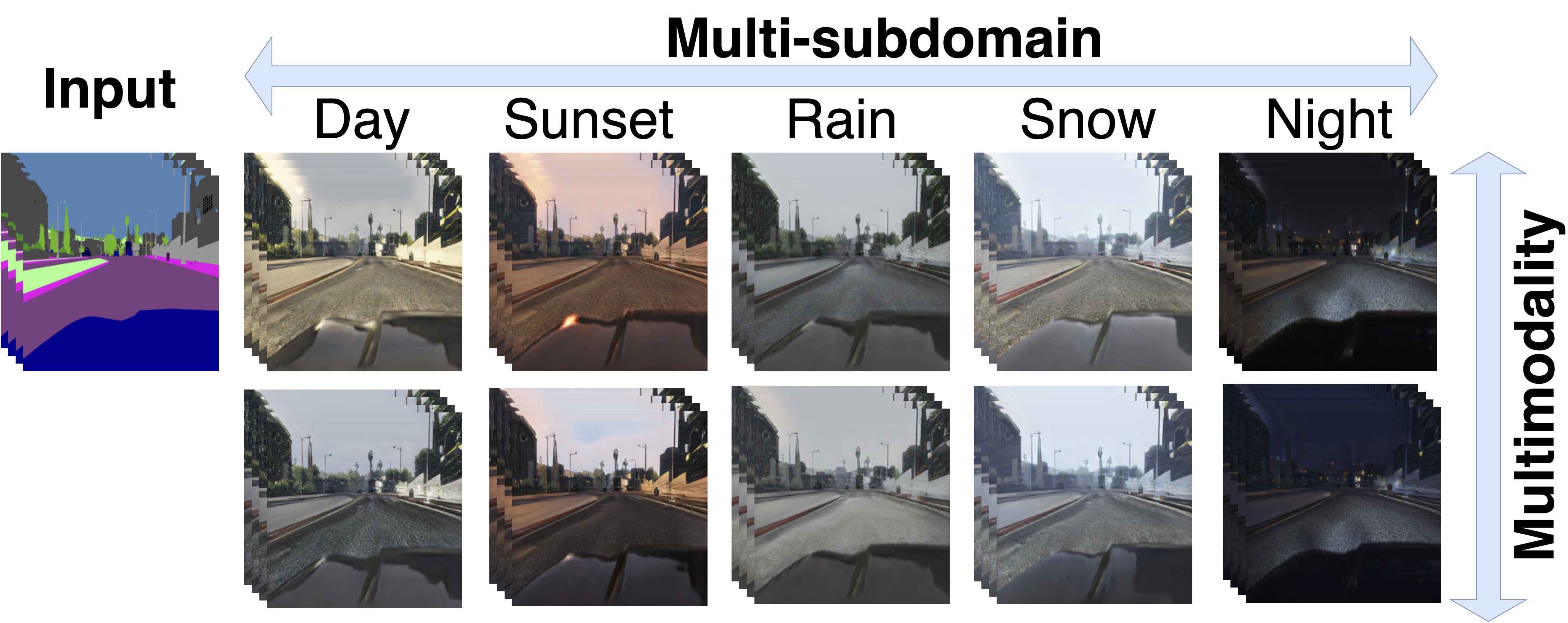}
\vspace{-0.5cm}
  \caption{Our proposed UVIT model can produce photo-realistic, spatio-temporal consistent translated videos in a multimodal way for multiple subdomains}
\label{fig:teaser}
\vspace{-0.5cm}
\end{figure}

In \SHORTTITLE{}, by assuming that all domains share the same underlying structure, namely content space, we exploit the style-conditioned translation. To simultaneously impose style and content consistency, we adopt an Encoder-RNN-Decoder architecture as the video translator, see~\fref{fig:overview} for an illustration of the proposed framework. There are two key ingredients in our framework:

\textbf{Conditional video translation:}  By applying the same style code to decode the content feature for a specific translated video, the translated video is style consistent. Besides, by changing the style code across videos, we achieve sub-domain\footnote{Hereafter we call it \textit{subdomain} and not domain because a subdomain must belong to a subset of a domain (for instance, subdomains of day, night, snow, etc. belong to the scene video domain)} and modality flexible video translation, see~\fref{fig:teaser} for an illustration of subdomains (columns) and modalities (rows). This overcomes the limitations of existing CycleGAN-based video translation techniques, \ie performing deterministic translations (generator as an injective function).

\textbf{Consistent video translation:}
Building inter-frame dependency is essential for generating dynamically vivid videos. Existing video translators utilize optical flow or 3D Convolutional Neural Networks (CNNs), which are incapable to fully capture the complex relationships between multiple video frames.
We adopts a RNN-based translator to incorporate inter-frame and current frame information in the high-dimensional hidden space from more frames. As validated by Bojanowski \emph{et al.} on image generation task~\cite{bojanowski2018optimizing}, integrating features in hidden space is beneficial to produce semantically meaningful, smooth nonlinear interpolation in image space.
Lacking paired supervision, another crucial aspect in unsupervised video translation lies in training criterion. Besides GAN~\cite{goodfellow2014generative} loss and spatial cycle consistency~\cite{zhu2017unpaired} loss, we propose video interpolation loss as the temporal constraint to strengthen semantic preserving. Specifically, we use translator building blocks to interpolate the current frame according to inter-frame information produced during translation. The current frame is then used as self-supervised target to tailor the inter-frame information. Meanwhile, it is validated that introducing self-supervision task is beneficial for cross-domain unsupervised tasks~\cite{sun2019unsupervised,chen2019self,ren2018cross}. Such self-supervision is applied to all building blocks of the translator, which stabilizes the challenging unpaired video adversarial learning. Effectiveness of this loss is validated through ablation study.

The main contributions of our paper are summarized as follows: 
\begin{enumerate}
    \item We introduce an Encoder-RNN-Decoder video translator, which decomposes the temporal consistency into independent style and content consistencies for more stable consistent video translation.
    
    \item Our style-conditioned decoder ensures style consistency as well as facilitates multimodal and multi-subdomain V2V translation.
    
    \item We use self-supervised learning to incorporate an innovative video interpolation loss which preserves inter-frame information according to the current frame. The combined translation frame is more semantically preserved w.r.t the corresponding input frame. Therefore, our RNN-based translator can recurrently generate dynamically vivid and photo-realistic video frames. 
    
\end{enumerate}

\section{Related Work}
\label{sec:relatedwork}

\smallskip
\noindent \textbf{Image-to-Image Translation.}
Most of the GAN-based I2I methods mainly focus on the case where paired data exists \cite{pix2pix2017,zhu2017toward,wang2018high}. However, with the cycle-consistency loss introduced in CycleGAN \cite{zhu2017unpaired}, promising performance has been achieved also for the unsupervised I2I~\cite{huang2018multimodal,almahairi2018augmented,liu2017unsupervised,mo2018instagan,romero2018smit,gong2019dlow,choi2018stargan,wu2019transgaga,cho2019image,wu2019relgan,alharbi2019latent}. 
The conditional distribution of the translated pictures on the input pictures is quite likely to be multimodal (\eg from a semantic label to different images in a fixed weather condition). However, traditional I2I problem often lacks this characteristic and produces an unimodal outcome. Zhu\ETAL{zhu2017toward} proposed Bicycle-GAN that can output diverse translations in a supervised manner. There are also some extensions~\cite{huang2018multimodal,almahairi2018augmented,karras2019style} of CycleGAN to decompose the style and content so that the output can be multimodal in the unsupervised scenario. Our work goes in this direction, and under the assumption that close frames within the same domain share the same style, we adopt the style control strategy in the image domain proposed by Almahairi\ETAL{almahairi2018augmented} to the video domain.

\smallskip
\smallskip
\noindent \textbf{Video-to-Video Translation}
In the seminal work, Wang\ETAL{wang2018video} (vid2vid) combined the optical flow and video-specific constraints and proposed a general solution for V2V in a supervised way, which achieves long-term high-resolution video sequences. However, vid2vid relies heavily on labeled data which makes it difficult to scale in unsupervised real-world scenarios. As our approach exploits the unsupervised V2V representation, it is the focus of our document.

Based on the I2I CycleGAN approach, recent methods~\cite{bashkirova2018unsupervised,bansal2018recycle,chen2019mocycle} on unsupervised V2V proposed to design spatio-temporal loss to achieve more temporally consistent results while preserving semantic information. Bashkirova\ETAL{bashkirova2018unsupervised} proposed a 3DCycleGAN method which adopts 3D convolutions in the generator and discriminator of the CycleGAN framework to capture temporal information. However, since the small 3D convolution operator (with a small temporal dimension 3) only captures dependency between adjacent frames. 3DCycleGAN therefore can not exploit temporal information for generating longer style consistent video sequences. Furthermore, the 3D discriminator is also limited in capturing complex temporal relationships between video frames. As a result, when the gap between input and target domain is large, 3DCycleGAN tends to sacrifice the image-level quality and generates blurry and gray translations.

Additionally, Bansal\ETAL{bansal2018recycle} designed a recycle loss (ReCycleGAN) for jointly modeling the spatio-temporal relationship between video frames and thus solving the semantic preserving problem. They trained a temporal predictor to predict the next frame based on two past frames, and plugged the temporal predictor in the cycle-loss to impose the spatio-temporal constraint on the traditional image-level translator. 
Although ReCycleGAN succeeds in V2V translation scenarios such as face-to-face or flower-to-flower, similar to CycleGAN, it lacks domain generalization as the translation fails to be consistent in domains with a large gap with respect to the input. We argue that there are two major reasons that affect ReCycleGAN performance in complex scenarios.
First, the translator is a traditional image-level translator without the ability to record the inter-frame information within videos. It processes input frames independently, which has limited capacity in exploiting temporal information, being not content consistent enough.
Second, ReCycleGAN temporal predictor only imposes the temporal constraint between a few adjacent frames, the generated video content still might shift abnormally: a sunny scene could change to a snowy scene in the following frames. 
Note that Chen\ETAL{chen2019mocycle} incorporate optical flow to add motion cycle consistency and motion translation constraints. However, their Motion-guided CycleGAN still suffers from the same two limitations as in ReCycleGAN.

In summary, previous methods fail to produce style consistent and multimodal video sequences. Besides, they lack the ability to achieve translation which is both content consistent enough and frame-wise realistic. In this paper, we propose \SHORTTITLE{}, a novel method for \TITLE{}, which produces high-quality semantic preserving frames with consistency within the video sequence. Besides, to the best of our knowledge, our method is the first method that jointly addresses multiple-subdomains and multimodality in V2V cross-domain translations.

\section{Unsupervised
Multimodal VIdeo-to-video Translation via Self-Supervised Learning (\SHORTTITLE{})}
\label{sec:UVIT}

\subsection{Problem setting}
Let $A$ be the video domain A, $a_{1:T} = \{a_1, a_2, ..., a_T \}$ be a sequence of video frames in $A$, let $B$ be the video domain B, $b_{1:T} = \{b_1, b_2, ..., b_T \} $ be a sequence of video frames in  $B$. For example, they can be sequences of semantic segmentation labels or scene images.  
Our general goal of unsupervised video-to-video translation is to train a generator to convert videos between domain A and domain B with many-to-many mappings. Either domain A or domain B can have multiple subdomains (sunny, snow, rain for the case of weather conditions).
More concretely, to generate the style consistent video sequence, we assume each video frame has a shared style latent variable $z$.
Let $z_a \in Z_A$ and $z_b \in Z_B$ be the style latent variables in domain A and B, respectively.

We aim to achieve two conditional video translation mappings: $\mathbb{G}_{AB}^{trans}: A\times Z_B \mapsto B^{trans}$ and $\mathbb{G}_{BA}^{trans}: B\times Z_A \mapsto A^{trans}$. 
As we propose to use the video interpolation loss to train the translator components in a self-supervised manner,
we also define the video interpolation mappings: $\mathbb{G}_A^{interp}: A\times Z_A \mapsto A^{interp}$ and $\mathbb{G}_B^{interp}: B\times Z_B \mapsto B^{interp}$. 
Interpolation and translation mappings use exactly the same building blocks.

\subsection{Translation and Interpolation pipeline}
\begin{figure*}[t]
\begin{center}
\includegraphics[width=0.8\linewidth]{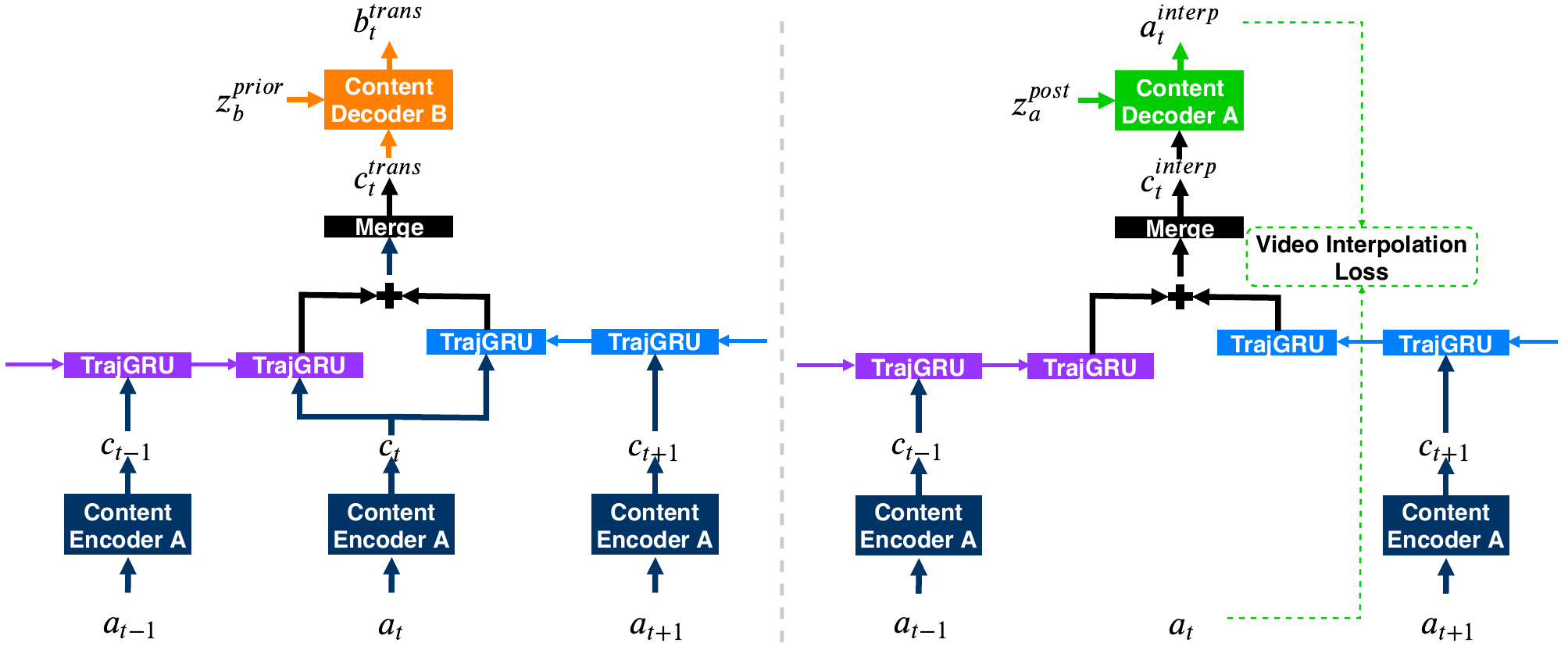}
\end{center}
\vspace{-0.4cm}
\caption{Video translation (left) and video interpolation (right): two processes share modules organically. The input latent content is processed by the Merge Module to merge information from TrajGRUs in both the forward and the backward direction. The translation content ($c^{trans}_{t}$) is obtained by updating interpolation content ($c^{interp}_{t}$) with the content ($c_{t}$) from the current frame ($a_t$)}
\label{fig:merge_module}
\vspace{-0.4cm}
\end{figure*}

In this work, inspired by UNIT~\cite{liu2017unsupervised}, we assume a shared content space such that
corresponding frames in two domains are mapped to the same latent content representation. We show the translation and interpolation processes in \fref{fig:merge_module}. To achieve the goal of unsupervised video-to-video translation, we propose an Encoder-RNN-Decoder translator which contains the following components:

\vspace{-1mm}
\begin{itemize}
    \item Two content encoders ($\mathbb{CE}_A$ and  $\mathbb{CE}_B$), which extract the frame-wise content information from each domain to the common spatial content space. 
    
    \item Two style encoders ($\mathbb{SE}_A$ and  $\mathbb{SE}_B$), which encode video frames to the respective style domains. 

    \item Two Trajectory Gated Recurrent Units (TrajGRUs) ~\cite{shi2017deep} to form a Bi-TrajGRU ($\mathbb{T}$), which propagates the inter-frame content information bidirectionally. TrajGRU~\cite{shi2017deep} is one variant of Convolutional RNN (Recurrent Neural Network)~\cite{xingjian2015convolutional}, which can actively learn the location-variant structure in the video data. More details in \AppendixName{}.
    \item One merge module ($\mathbb{M}$), which adaptively combines the inter-frame content from two directions. 
    \item Two conditional content decoders ($\mathbb{CD}_A$ and $\mathbb{CD}_B$), which take the spatio-temporal content information and the style code to generate the output frame. If needed, it also takes the conditional subdomain information as an one hot vector encoding.
\end{itemize} 
\vspace{-0.3cm}

\smallskip
\noindent \textbf{Video translation:}
Given an input frame sequence ($..., a_{t-1}, a_{t}, a_{t+1},...$), we extract the posterior style ($z_{a}^{post}$) from the first frame ($a_{1}$) with a style encoder ($\mathbb{SE}_A$). Additionally, we extract each content representation ($...,c_{t-1}, c_{t}, c_{t+1},...$) with the content encoder ($\mathbb{CE}_A$). 

Translation is conducted in a recurrent way. 
To get the translation result $b_{t}^{trans}$ for time $t$ , we process the independent content representation: (1) propagate content for the surrounding frames ($...,c_{t-1}, c_{t+1},...$) through \textit{Bi-TrajGRU} ($\mathbb{T}$) to obtain the inter-frame content information. (2) update this information with the current frame content ($c_{t}$) (see \fref{fig:merge_module} left, Merge Module $\mathbb{M}$) to get the spatio-temporal content ($c_{t}^{trans}$) for translation.
At last, using the same style-conditioned strategy as Augment CycleGAN \cite{almahairi2018augmented,dumoulin2016learned,perez2018film}, the content decoder ($\mathbb{CD}_B$) takes the prior style information ($z_{b}^{prior}$) as the condition and utilizes $c_{t}^{trans}$ to generate the translation result ($\mathbb{CD}_B(c_{t}^{trans},z_{b}^{prior})= b_{t}^{trans}$). This process is repeated until we get the whole translated sequence (...,$b_{t-1}^{trans}$, $b_{t}^{trans}$,$b_{t+1}^{trans}$,...). 


Style code is induced as the condition of (AdaIN-based \cite{huang2017arbitraryadain}) content decoder. If a domain (\emph{e.g.} scene images) is presorted, we have prior information on which subset (rain, night, etc.) a video belongs to, so we can take such prior information as a subdomain (subset) label to achieve deterministic control for the style. Within each subset, there are still different modalities (\emph{e.g.} overcast day, sunny day in day subdomain), yet we do not have prior access to it. This modality information is therefore learned by style encoder. Subdomain label (taken as one-hot vector if available) and modality information together constitute 21-dimensional style code. Style consistency is ensured by sharing style code among a specific video sequence. Multimodal translation is realized by inducing different style codes across videos. When subdomain information is unavailable, simply using style encoder to learn subdomain styles as modalities, we can still generate multimodal style consistent results in a stochastic way.

\smallskip
\noindent \textbf{Video interpolation:} 
In video translation process in \fref{fig:merge_module}, when translating a specific frame ($a_t$), the translation content  ($c_t^{trans}$) is integrated by the current frame content ($c_t$) and inter-frame information ($c_t^{interp}$) from the surrounding frames ($...,a_{t-1},a_{t+1},...$). The inter-frame content information helps to build up the dependency between each frame and its surrounding frames, ensuring content consistency across frames. However, if $c_t^{interp}$ and $c_t$ are not aligned well, image-level quality can be affected. The translated frame ($b_t^{trans}$) will incline to over smooth image-level details and sacrifice high-level semantic correspondence with $a_t$. Tailoring inter-frame information is thus of pivotal importance.

Thanks to the flexible Encoder-Decoder structure, our decoder can generate interpolated frame ($a_t^{interp}$) from $c_t^{interp}$. Video interpolation loss is proposed to compute the L1-Norm distance between interpolated frame($a_t^{interp}$) and the current frame($a_t$), which adds supervision to the inter-frame content($c_t^{interp}$). $c_t^{interp}$ can thus be combined with $c_t$ more organically. Therefore, the translation task directly benefits from the interpolation task, producing more semantic preserving and photo-realistic outcomes. 

Meanwhile, such self-supervised training would be beneficial to make the network more stable in the challenging unpaired video adversarial learning \cite{sun2019unsupervised}.  GANs are powerful methods to learn a data probability distribution with no supervision, yet training GANs is well known for being delicate, unstable~\cite{arjovsky1701towards,mao2017least,arjovsky2017wasserstein} and easy to suffer from mode collapse\cite{bansal2018recycle}. Besides cycle loss acting as spatial constraint, we introduce the video interpolation loss as a temporal constraint for GAN training in a self-supervised way. It have been validated that bringing self-supervision is beneficial for cross-domain unsupervised tasks (\emph{e.g.}, natural image synthesis)~\cite{sun2019unsupervised,chen2019self,ren2018cross}.

Furthermore, our framework aims to learn latent representation for style and content, while it has been empirically observed~\cite{gulrajani2016pixelvae,chen2016variational} that it is non-trivial to
use latent variables when coupled with a strong autoregressive decoder (\emph{e.g.}, RNN). Goyal\ETAL{goyal2017z} found that auxiliary cost could ease training of the latent variables in RNN-based generative latent variable models. Therefore, the video interpolation task provides the latent variables with a auxiliary objective that enhances the performance of the overall model.

Note that the proposed temporal loss highly differs from the previous ReCycleGAN loss~\cite{bansal2018recycle} as: \textit{(1)} we use a RNN-based architecture that captures temporal information better in a high-level feature space, \textit{(2)} interpolation is conducted within the translator building blocks rather than using different modules, training the translator with direct self-supervision, \textit{(3)} the translator directly utilizes tailored inter-frame information for better semantic preserving translations.

\subsection{Loss functions}
We use the Relativistic GAN (RGAN)~\cite{jolicoeur2018relativistic} and the least square~\cite{mao2017least} version for the adversarial loss. RGAN estimates the probability that the given real data is more realistic than a randomly sampled fake data. 

We use image-level discriminators ($D_x^{img}$) and video-level ($D_x^{vid}$) discriminators to ensure that output frames resemble a real video clip in both video-level and image-level. Moreover, we also add style discriminators ($D_{Z_{x}}$) to adopt an adversarial approach for training style encoders. 

\smallskip
\noindent \textbf{Video adversarial loss.}
The translated video frames aim to be realistic compared to the real samples in the target domain for both an image-level and a video-level basis.
\begin{equation}
      L_B^{adv} =  \frac{1}{T} \sum_{i=1}^{i=T}[ D_B^{img}(b_{i}^{trans}) - D_B^{img}(b_i) - 1]^2 
      +  [ D_B^{vid}(b_{1:T}^{trans}) - D_B^{vid}(b_{1:T}) - 1]^2, 
\end{equation}
where, $b_{1:T}^{trans}$ are the translated frames from time $1$ to $T$.
$D_B^{img}$ is the image-level discriminator for domain B , $D_B^{vid}$ is the video-level discriminator for domain B. Adversarial loss for domain A ($ L_{A}^{adv}$) is defined similarly.
 
\smallskip
\noindent \textbf{Video interpolation loss.}
The interpolated video frames should be close to the ground truth frames (pixel-wise loss). Additionally, they aim to be realistic compared to other real frames within the domain (adversarial loss). 
\begin{equation*}
      L_{A}^{interp}\!=\!\frac{1}{(T\!-\!2)}  (\lambda_{interp}\!\!\parallel a_{2:T-1}-a_{2:T-1}^{interp} \parallel_{1}\!+\!\!\!\!\sum_{i=2}^{i=T-1}\!\!\![ D_A^{img}(a_{i}^{interp}) - D_A^{img}(a_i) - 1]^2). 
\end{equation*}
Since we are using bidirectional TrajGRUs, we use frames from time $2$ to $T-1$ to compute the video interpolation loss. $a_{2:t-1}^{interp}$ are the interpolated frames. The first part of the loss is the supervised pixel-wise $L_1$ loss, and the later part is the GAN loss computed on the image-level discriminator $D_A^{img}$. $\lambda_{interp}$ is used to control the weight between two loss elements. 

\smallskip
\noindent \textbf{Cycle consistency loss.} 
In order to ensure semantic consistency in an unpaired setting, we use a cycle-consistency loss:

\begin{equation}
\begin{aligned}
      L_{A}^{cycle} = \frac{\lambda_{cycle}}{T} \parallel a_{1:T}-a_{1:T}^{rec} \parallel_{1},
\end{aligned}
\end{equation} 
where
$a_{1:T}^{rec}$ are the reconstructed frames of domain A from time $1$ to $T$, \emph{i.e. } $a_{1:T}^{rec}= \mathbb{G}_{BA}^{trans}(b_{1:T}^{trans}, z_a^{post})$. 
Where $z_a^{post}$ is the posterior style variable produced by using the style encoder to encode $a_{1}$. $\lambda_{cycle}$ is the cycle consistency loss weight. 

\smallskip
\noindent \textbf{Style encoder loss.} To train the style encoder, the style reconstruction loss and style adversarial loss are define in a similar way as Augment CycleGAN~\cite{almahairi2018augmented}: 
\begin{equation}
      L_{Z_A}^{style} = \lambda_{rec}\parallel z_a^{rec} - z_a^{prior} \parallel_{1} +  [ D_{Z_A}(z_a^{post}) - D_{Z_A}(z_a^{prior}) - 1]^2.
\end{equation} 
 Here,  $z_a^{prior}$ is the prior style latent variable of domain A drawn from the prior distribution. $z_a^{rec}$ is the reconstructed style latent variable of domain A by using the style encoder to encode $a_1^{trans}$. $\lambda_{rec}$ is the style reconstruction loss weight. 
 
Therefore, the objective for the generator is:
 
\begin{equation}
     L_G^{total} = L_A^{adv} + L_B^{adv} + L_A^{interp} + L_B^{interp} +
     L_A^{cycle} + L_B^{cycle} + L_{Z_A}^{style} + L_{Z_B}^{style}
\end{equation}
Detailed $\lambda$ values and loss functions for discriminators are attached in the \AppendixName{}. Detailed training algorithm for RGANs can be found in~\cite{jolicoeur2018relativistic}.

\section{Experiments}
\label{sec:experiments}

We validate our method using two common yet challenging datasets: Viper~\cite{richter2017playing}, and Cityscapes~\cite{cordts2016cityscapes} datasets. In order to feed more frames within limited single GPU resource, we use the image with $128 \times 128$ and 10 frames per batch for the main experiments. During inference, we use video sequences of 30 frames. These 30 frames are divided into 4 smaller sub-sequences of 10 frames with overlap. They all share the same style code to be style consistent.  Note that our model can be easily extended to process longer style-consistent video sequences by keeping sharing the same style code for the sub-sequences. The video example of longer style consistent video is provided in \AppendixName{}, where detailed description of the dataset and implementation are also attached.

\subsection{Ablation Study}
\label{sec:experimentsablation}

In order to demonstrate the contribution of our method, we first conduct ablation study experiments. We provide quantitative and qualitative experimental results that evidence the proposed video interpolation loss for a better V2V translation. Besides, we study how the number of frames influence the semantic preserving performance. We also provide multimodal consistent results of our model trained without using subdomain label in the \AppendixName{}.

\smallskip
\noindent \textbf{Video interpolation loss.} 
We provide ablation experiments to show the effectiveness of the proposed video interpolation loss. 
We conduct experiments on both the image-to-label and the label-to-image tasks. We denote \SHORTTITLE{} trained without video interpolation loss as ''\SHORTTITLE{} wo/vi''.

We follow the experimental setting of ReCycleGAN~\cite{bansal2018recycle} and use semantic segmentation metrics to quantitatively evaluate the image-to-label results.
The Mean Intersection over Union (mIoU), Average Class Accuracy (AC) and Pixel Accuracy (PA) scores for ablation experiments are reported in \tref{table:comparison_ablation}. Our model with video interpolation loss achieves the best performance across subdomains, which confirms that the video interpolation helps to preserve the semantic information between the translated frame and the corresponding input frame.

\begin{table}[t]
\centering
 \caption{\textbf{Image-to-Label (Semantic segmentation) quantitative evaluation.} We validate UVIT without video interpolation loss ($wo/vi$) under Mean Intersection over Union (mIoU), Average Class Accuracy (AC) and Pixel Accuracy (PA) scores}
 \begin{tabular}{c c c c c c c }\toprule
  Criterion & Model & Day & Sunset & Rain & Snow & Night\\\midrule
\multirow{2}*{mIoU $\uparrow$}
&\SHORTTITLE{} wo/vi & 10.14&	10.70&	11.06 &	10.30&	9.06\\
  ~&\SHORTTITLE{}&\textbf{13.71} &\textbf{13.89} & \textbf{14.34}&\textbf{13.23}&\textbf{10.10} \\
  \midrule
  
 \multirow{2}*{AC $\uparrow$}&\SHORTTITLE{} wo/vi &15.07&	15.78&	15.46&	15.01&	13.06\\
    ~&\SHORTTITLE{} &\textbf{18.74} &\textbf{19.13} & \textbf{18.98}& \textbf{17.81}	&\textbf{13.99}\\  \midrule   
  
    \multirow{2}*{PA $\uparrow$}&\SHORTTITLE{} wo/vi &56.33&	57.16&	58.76&	55.45&	55.19\\
    ~&\SHORTTITLE{} &  \textbf{68.06}& \textbf{66.35}& \textbf{67.21}&	\textbf{65.49}&	\textbf{58.97} \\

  \bottomrule
 \end{tabular}
  \vspace{-0.25cm}
 \label{table:comparison_ablation}
\end{table}

\begin{table}[t]
\centering
 \caption{\textbf{Label-to-image quantitative evaluation.} We validate our system without video interpolation loss ($wo/vi$) under the Fr\'echet Inception Distance (FID) score}  
 \begin{tabular}{c c c c c c c}\toprule
  Criterion & Model & Day & Sunset & Rain & Snow & Night\\\midrule
  \multirow{2}*{FID $\downarrow$}&\SHORTTITLE{} wo/vi & 26.95&23.04& 30.48&34.62& 47.50\\
    ~&\SHORTTITLE{}&\textbf{17.32}&\textbf{16.79}&\textbf{19.52}&\textbf{18.91}& \textbf{19.93}\\   

  \bottomrule
 \end{tabular}
 \label{table:comparison_FID_2}
 \vspace{-0.4cm}
\end{table}

For the label-to-image task, we use the Fr\'echet Inception Distance (FID)~\cite{heusel2017gans} to evaluate the feature distribution distance between translated videos and ground-truth videos. Similar to vid2vid~\cite{wang2018video}, we use the pre-trained network (I3D~\cite{carreira2017quo}) to extract features from videos.
We extract the semantic labels from the respective sub-domains to generate videos and evaluate the FID score on all the subdomains of the Viper dataset.
\tref{table:comparison_FID_2} shows the FID score for \SHORTTITLE{} and the corresponding ablation experiment. On both the image-to-label and label-to-image tasks, the proposed video interpolation loss plays a crucial role for \SHORTTITLE{} to achieve good translation results. 

\smallskip
\noindent \textbf{Different number of input frame.} Our RNN-based translator incorporates temporal information from multiple frames. We also investigate the influence of frame number on the performance of our model. As shown in \tref{Ablation2} \SHORTTITLE{} can achieve better semantic preserving with more frames feeding during training as the RNNs are better trained to leverage the temporal information. Specifically, for the image-to-label translation, with the increase of the number from 4 to 10, the overall mIoU increase from 11.19 to 13.07. Complete table and analysis are attached in \AppendixName{}.

\begin{table*}[t]
\centering
  \caption{\textbf{Quantitative results of  \SHORTTITLE{} with different number of frames per batch in training on the image-to-label (Semantic segmentation) task.} With the increase of input frames number in the sub-sequence, our RNN-based translator can utilize the temporal information better, resulting in better semantic preserving}
 \begin{tabular}{c c c c c c c c}\toprule
  Criterion&Frame number&Day&Sunset&Rain&Snow&Night&All\\\midrule
  \multirow{4}*{mIoU$\uparrow$}&4 &11.84&	11.91&	12.35&	11.37&	8.49&	11.19\\
  ~&6 &{12.29}&12.66&{13.03}&{11.77}&	{9.79}&{11.94}\\
  ~& 8 &{13.05}&13.21&14.23&{13.07}&	\textbf{11.00} &{12.87}\\
  ~& 10 &\textbf{13.71} &\textbf{13.89} & \textbf{14.34}&\textbf{13.23}&10.10	&\textbf{13.07} \\
  \bottomrule
 \end{tabular}
  \label{Ablation2}
\vspace{-0.3cm}
\end{table*}

\subsection{Comparison of \SHORTTITLE{} with State-of-the-Art Methods}
\label{sec:experimentsother}

\smallskip
\noindent \textbf{Image-to-label mapping.} To further ensure reproducibility, we use the same setting as our ablation study to compare \SHORTTITLE{} with ReCycleGAN~\cite{bansal2018recycle} in the image-to-label mapping task.
We report the mIoU, AC and PA metrics by the proposed approach and competing methods in \tref{table:comparison}. The results clearly validate the advantage of our method over the competing approaches in terms of preserving semantic information.  Our model can effectively leverage the inter-frame information from more frames in a direct way, which utilizes the temporal information better than the indirect way in ReCycleGAN~\cite{bansal2018recycle}.

\begin{table*}[t]
\centering
  \caption{\textbf{Quantitative comparison between \SHORTTITLE{} and baseline approaches on the image-to-label (Semantic segmentation) task.} Our translator effectively leverage the temporal information directly, thus producing more semantic persevering translation outcomes} 
 \begin{tabular}{c c c c c c c c}\toprule
  Criterion&Model&Day&Sunset&Rain&Snow&Night&All\\\midrule
  \multirow{5}*{mIoU$\uparrow$}
  &Cycle-GAN&3.39&3.82&3.02&3.05&7.76&4.10\\
  ~&ReCycleGAN (Reproduced)\footnotemark[1]&10.31&	11.18&	11.26&	9.81&	7.74&	10.11\\
  ~&ReCycleGAN (Reported)\footnotemark[2] &8.50&13.20&10.10&9.60&3.10&8.90\\
 ~& \SHORTTITLE{} (Ours) &\textbf{13.71} &\textbf{13.89} & \textbf{14.34}&\textbf{13.23}&\textbf{10.10}	&\textbf{13.07} \\
  \midrule 
  
   \multirow{5}*{AC$\uparrow$}
   &Cycle-GAN&7.83&8.56&7.91&7.53&11.12&8.55\\
   ~&ReCycleGAN (Reproduced)\footnotemark[1] &15.78&	15.80&	15.95&	15.56&	11.46&	14.84\\
   ~&ReCycleGAN (Reported)\footnotemark[2] &12.60&13.20&10.10&13.30&5.90&12.40\\
   ~& \SHORTTITLE{} (Ours) &\textbf{18.74} &\textbf{19.13} & \textbf{18.98}& \textbf{17.81}	&\textbf{13.99}& \textbf{17.59}\\
  \midrule
  
    \multirow{5}*{PA$\uparrow$}
    &Cycle-GAN&15.46&	16.34&	12.83&	13.20&	49.03&	19.59\\
    ~&ReCycleGAN (Reproduced)\footnotemark[1]&54.68&	55.91&	57.72&	50.84&	49.10&	53.65\\
    ~&ReCycleGAN (Reported)\footnotemark[2] &48.70&\textbf{70.00}&60.10&58.90&33.70&53.70 \\
     ~& \SHORTTITLE{} &  \textbf{68.06}& {66.35}& \textbf{67.21}&	\textbf{65.49}&	\textbf{58.97}& \textbf{65.20}\\
  
  \bottomrule
 \end{tabular}
 \label{table:comparison}
    \vspace{-0.3cm}
\end{table*}

\begin{table}[t]
\centering
  \caption{\textbf{Quantitative comparison between \SHORTTITLE{} and baseline approaches on the label-to-image task.} Better FID indicates that our translation has better visual quality and temporal consistency}
 \begin{tabular}{c c c c c c c}\toprule
  Criterion&Model&Day&Sunset&Rain&Snow&Night\\\midrule
  \multirow{3}*{FID$\downarrow$}
  &ReCycleGAN~\cite{bansal2018recycle} &23.60 &24.45 &28.54 &31.58 & 35.74 \\
  ~&Improved ReCycleGAN &20.39 &21.32 &25.67 &21.44& 21.45 \\
  ~&\SHORTTITLE{} (ours)&\textbf{17.32}&\textbf{16.79}&\textbf{19.52}&\textbf{18.91}& \textbf{19.93} \\
  \bottomrule
 \end{tabular}

   \label{table:comparison_FID}
  \vspace{-0.25cm}
\end{table}

\begin{table}[t]
\centering
 \caption{\textbf{Label-to-image Human Preference Score.} Our method outperforms all the competing unsupervised methods in both video-level and image-level reality. Note that we achieve comparable performance with vid2vid although it is supervised}
 \begin{tabular}{c c c}\toprule
      Human Preference Score & Video level & Image level\\\midrule
  \SHORTTITLE{} (ours) / Improved ReCycleGAN& \textbf{0.67} / 0.33& \textbf{0.66} / 0.34\\
  \SHORTTITLE{} (ours) / 3DCycleGAN~\cite{bashkirova2018unsupervised}& \textbf{0.75} / 0.25& \textbf{0.70} / 0.30\\
  \SHORTTITLE{} (ours) / vid2vid~\cite{wang2018video}& 0.49 / \textbf{0.51}&--\\
\SHORTTITLE{} (ours) / CycleGAN~\cite{zhu2017unpaired}& -- &\textbf{0.61} / 0.39\\
  \bottomrule
 \end{tabular}
 \label{table:HPS}
 \vspace{-0.4cm}
\end{table}

\smallskip
\noindent \textbf{Label-to-image mapping.} In this setting, we compare the quality of the translated video sequence by different methods.
We first report the FID score~\cite{heusel2017gans} on all the sub-domains of the Viper dataset in the same setting as our ablation experiments.
As the original ReCycleGAN output video sequences  can not ensure style consistency, just as shown in \fref{fig:compareHR},
we also report the results achieved by our improved version of the ReCycleGAN for a fair comparison.
Concretely, we develop a conditional version which formally controls the style of generated video sequences in a similar way as our \SHORTTITLE{} model, and denote the conditional version as improved ReCycleGAN.
The FID results by different methods are shown in \tref{table:comparison_FID}.
The proposed \SHORTTITLE{} achieves better FID on all the 5 sub-domains, which validates the effectiveness of our model in achieving better visual quality and temporal consistency.
Combining \tref{table:comparison_FID_2} and \tref{table:comparison_FID}, there is another observation -- the \SHORTTITLE{} w/o vi-loss could not dominate the Improved ReCycleGAN in terms of FID. This shows that the video interpolation loss is crucial for the superiority of our spatio-temporal translator.
\footnotetext[1]{ The result is reproduced by us. The output would be in a resolution of $256 \times 256$, we then downscale it to $128 \times 128$ to compute the statistics.}
\footnotetext[2]{ This is the result reported in the original paper~\cite{bansal2018recycle} with a resolution of $256 \times 256$.}

\begin{figure}
\vspace{-0.4cm}
    \centering
    \includegraphics[scale=0.99]{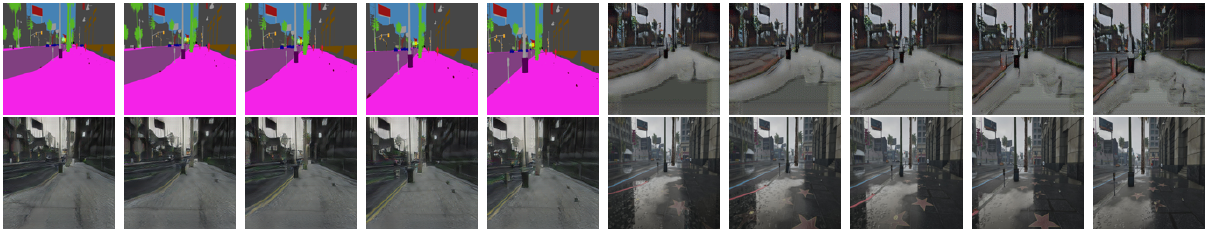}
    \vspace{-0.3cm}   
    \caption{\textbf{Label-to-image qualitative comparison.} Top left: label inputs; Top right: improved ReCycleGAN outputs; Bottom left: \SHORTTITLE{} outputs. Bottom right: ground truth. Video examples can be found in \AppendixName{}}
    \label{fig:compare1}
    \vspace{-1.2cm}
\end{figure}

To thoroughly evaluate the visual quality of the video translation results,
we conduct a subjective evaluation on the Amazon Mechanical Turk (AMT) platform. The detailed information of conducting this subjective test is provided in the \AppendixName{}.
We compare the proposed \SHORTTITLE{} with 3DCycleGAN and ReCycleGAN. The video-level and image-level human preference scores (HPS) are reported in \tref{table:HPS}.
For reference, we also compare the video-level quality between \SHORTTITLE{} and the supervised vid2vid model~\cite{wang2018video}. Meanwhile, image-level quality comparison between \SHORTTITLE{} and CycleGAN (the image translation baseline) is also included.
\tref{table:HPS} clearly demonstrates the effectiveness of our proposed \SHORTTITLE{} model.
In the video-level comparison, our unsupervised \SHORTTITLE{} model outperforms the competing unsupervised ReCycleGAN and 3DCycleGAN by a large margin, and achieves comparable results with the supervised benchmark.
In the image-level comparison, \SHORTTITLE{} achieves better HPS than both the V2V competing approaches and the image-to-image baseline. 
Qualitative examples in \fref{fig:compare1} also show that \SHORTTITLE{} model produces a more content consistent video sequence. It could not be achieved by simply introducing the style control without the specialized network structure to record the inter-frame information. For a better comparison, we include several examples of generated videos in the \AppendixName{}.

\subsection{More experimental results}

\begin{figure}
\vspace{-0.6cm}
    \centering
    
    \includegraphics[scale=1]{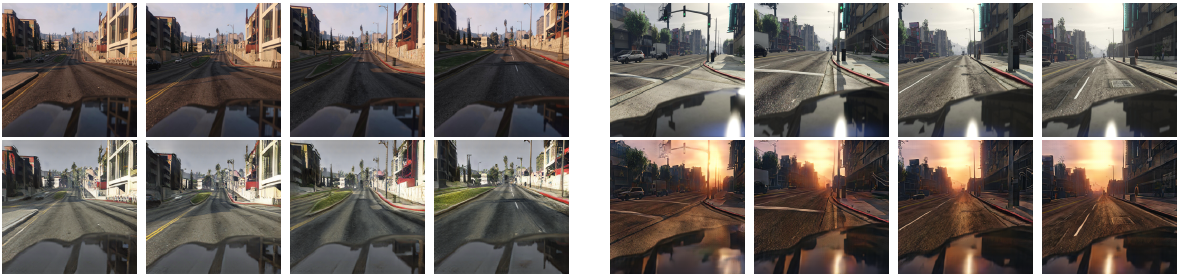}
\vspace{-0.3cm}
    \caption{\textbf{Viper Sunset-and-Day} Top left: input Sunset video; Top right: input Day video; Bottom left:  translated Day video; Bottom right: translated Sunset video}
    \label{fig:s2dHD}
    \vspace{-0.5cm}
\end{figure}

\noindent \textbf{High resolution results.} To get a higher resolution and show more details within the existing GPU constraint, we also train our model using images of $256 \times 256$ and 4 frames per batch,
then test with longer sequence, which is divided into subsequences of 4 frames with overlap. 
A visual example is shown in \fref{fig:compareHR}. More results and videos are provided in \AppendixName{}. 

\noindent \textbf{Translation on other datasets.} Besides translating video sequences between image and semantic label domains, we also train models to translate video sequences between different scene image subdomains and different video datasets.

\begin{figure}
\vspace{-0.3cm}
    \centering

    \includegraphics[scale=1]{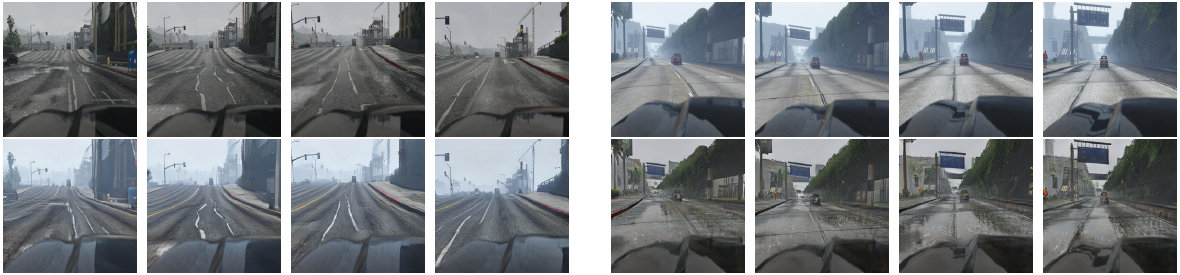}
\vspace{-0.3cm}
   \caption{\textbf{Viper Rain-and-Snow} Top left: input Rain video; Top right: input Snow video; Bottom left: translated Snow video; Bottom right: translated Rain video}
    \label{fig:r2sHD}
    \vspace{-0.3cm}
\end{figure}

\begin{figure}
\vspace{-0.7cm}
    \centering
    
    \includegraphics[scale=1]{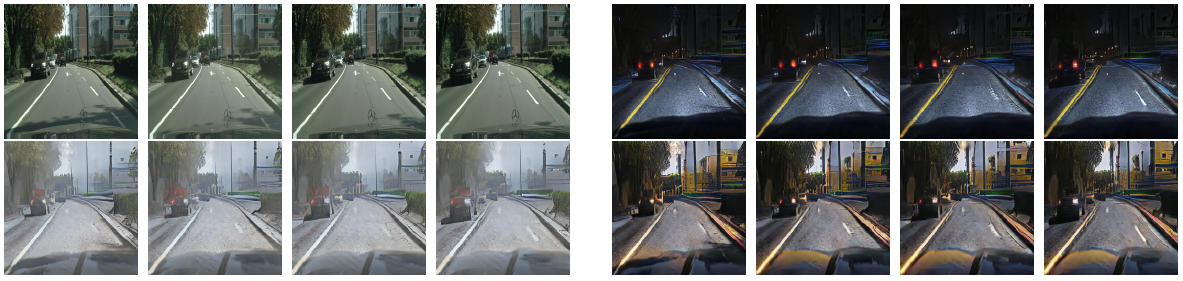}
\vspace{-0.3cm}    
    \caption{\textbf{Cityscapes to Viper translation.} Top left: input Cityscapes video; Top right: translated Viper video in the night scenario; Bottom left: translated Viper video in the snow scenario; Bottom right: translated Viper video in the sunset scenario}
    \label{fig:c2g}
     \vspace{-0.7cm}
\end{figure}

In \fref{fig:s2dHD} and \fref{fig:r2sHD}, we provide visual examples of video translation of  Sunset-and-Day and Rain-and-Snow scenes in the Viper dataset.
Visual examples of translation between Viper and Cityscapes~\cite{cordts2016cityscapes} datasets is organized in figure \ref{fig:c2g}.
They show the ability of our approach to learn the association between synthetic videos and real-world videos. 
More examples and the corresponding videos are attached in \AppendixName{}.

\section{Conclusion}

In this paper, we have proposed UVIT, a novel method for unsupervised video-to-video translation. 
A specialized Encoder-RNN-Decoder spatio-temporal translator has been proposed to decompose style and content in the video for temporally consistent and modality flexible video-to-video translation.
In addition, we have designed a video interpolation loss within the translator which utilizes highly structured video data to train our translators in a self-supervised manner. This enables the effective application of RNN-based network in the challenging V2V task.
Extensive experiments have been conducted to show the effectiveness of the proposed UVIT model.
Without using any paired training data, the proposed UVIT model is capable of producing excellent multimodal video translation results, which are image-level realistic, semantic information preserving and video-level consistent.
\\
\\
\noindent\textbf{Acknowledgments. }
This work was partly supported by the ETH Z\"urich Fund (OK), and by Huawei, Amazon AWS and Nvidia grants.

\clearpage
%
%
\bibliographystyle{splncs04}
\bibliography{egbib2}
\clearpage


\setcounter{section}{0}

\noindent \textbf{\textit{{\Large{Supplementary Material}}}} 
\bigskip

In this supplementary material, we provide more details and results of our method as follows:

\begin{enumerate}
    \item Additional loss details and implementation details;
    \item More experimental results with the resolution of $128 \times 128$; 
    \item More experimental results with the resolution of $256 \times 256$; 
    \begin{itemize}
        \item Translation between image and label: comparison with the baseline, multimodal results and a 1680-frame style-consistent sequence;
        \item Translation on other datasets: Rain and Snow, Sunset and Day, Viper and Cityscapes;
    \end{itemize}
\item Additional ablation experiments details.
\end{enumerate}

All the videos can be found in our project website:

\href{https://uvit.netlify.com/}{Project Website https://uvit.netlify.com/}

\section{Additional Loss Details and Implementation Details}
\subsection{Loss functions for the discriminator}

In this section, we provide more details of our image-level ($D^{img}$), video-level ($D^{vid}$), and style latent ($D_{Z}$) discriminator losses. 
For the purpose of simplicity, we only present the loss functions for domain A, and the loss functions  for domain B are defined following the same set of equations. 
Our adversarial loss is based on Relativistic GAN (RGAN) \cite{jolicoeur2018relativistic}, which tries to predict the probability that a real sample is relatively more realistic than a fake one. 

\paragraph{Image level discriminator loss} The loss term $D_A^{img}$ is defined as follows: 
\begin{equation}
\begin{aligned}
      L_{D_A^{img}}^{GAN} =
      \frac{1}{2(T-2)}\sum_{i=2}^{i=T-1}[ D_A^{img}(a_{i}) - D_A^{img}(a_i^{interp}) - 1]^2 \\
     + \frac{1}{2T}\sum_{i=1}^{i=T}[ D_A^{img}(a_i) - D_A^{img}(a_{i}^{trans}) - 1]^2.
\end{aligned}
\end{equation}

\paragraph{Video level discriminator loss} $D_A^{vid}$ for domain A is defined as follows:
\begin{equation}
\begin{aligned}
      L_{D_A^{vid}}^{GAN} =
      [D_A^{vid}(a_{1:T}) - D_A^{vid}(a_{1:T}^{trans}) - 1]^2. 
\end{aligned}
\end{equation}

\paragraph{Style latent variable discriminator loss} This loss term ($D_{Z_A}$) for the style domain A is defined as follows: 
\begin{equation}
\begin{aligned}
      L_{D_{Z_A}}^{GAN} = [ D_{Z_A}(z_a^{prior}) - D_{Z_A}(z_a^{post}) - 1]^2.
\end{aligned}
\end{equation}

\subsection{Network structure}
\paragraph{Style Encoder, Content Encoder and Content Decoder}

Our style encoder is similar to the one used in Augment CycleGAN \cite{almahairi2018augmented}. Under the shared content space assumption \cite{liu2017unsupervised}, we decompose the style-conditioned Resnet-Generator  used in Augment CycleGAN \cite{almahairi2018augmented} into a Content Encoder and a Content Decoder. 
Moreover, when the sub-domain information is available, we assign part of the style latent variable to record such prior information.
Concretely, we use one-hot vector to encode the sub-domain information. 

\paragraph{RNN - Trajectory Gated Recurrent Units (TrajGRUs)} 
Traditional RNN (Recurrent Neural Network) is based on the fully connected layer, which has limited capacity of profiting from the underlying spatio-temporal information in video sequence.
In order to take full advantage of the spatial and temporal correlations, UVIT utilizes a convolutional RNN architecture in the generator.
TrajGRU~\cite{shi2017deep} is one variant of Convolutional RNN (Recurrent Neural Network)~\cite{xingjian2015convolutional}, which can actively learn the location-variant structure in the video data. 
It uses the input and hidden state to generate the local neighborhood set for each location at each time, thus warping the previous state to compensate for the motion information. We take two TrajGRUs to propagate the inter-frame information in both directions in the shared content space.

\paragraph{Discriminators ($D^{img}, D^{vid}, D_{Z}$)}
For the image-level discriminators $D^{img}$, the architecture is based on the PatchGANs~\cite{pix2pix2017} approach. Likewise, Video-level discriminators $D^{vid}$ are similar to PatchGANs, yet we employ 3D convolutional filters. For the style latent variable discriminators $D_{Z}$, we use the same architecture as in Augmented CycleGAN~\cite{almahairi2018augmented}.

\subsection{Datasets} 
We validate our method using two common yet challenging datasets: Viper~\cite{richter2017playing}, and Cityscapes~\cite{cordts2016cityscapes} datasets.

\textbf{Viper} has semantic label videos and scene image videos. There are 5 subdomains for the scene videos: day, sunset, rain, snow and night. The large diversity of scene scenarios makes this dataset a very challenging testing bed for the unsupervised V2V task. We quantitatively evaluate translation performance by different methods on the image-to-label and the label-to-image mapping tasks. We further conduct the translation between different subdomains of the scene videos for qualitative analysis. 

\textbf{Cityscapes} has real-world street scene videos. As there is not subdomain information for Cityscapes, we conduct experiments without subdomain label for Cityscapes. We conduct qualitative analysis on the translation between scene videos of Cityscapes and Viper dataset.  Note that there is no ground truth semantic labels for the continuous Cityscapes video sequences. The semantic labels are only available to a limited portion of none-continuous individual images. Therefore, we could not use it for our evaluation of image-to-label (semantic segmentation) performance.

\subsection{Implementation Details}
We train our model using images of $128 \times 128$ and 10 frames per batch in a single NVIDIA P100 GPU for the main experiments to capture temporal information with more frames. Setting the batch size to one, it takes about one week to train. Note that it takes roughly 4 days to train using 6 frames per batch. 

During inference, we use video sequences of 30 frames.
These 30 frames are divided into 4 smaller sequences of 10 frames with overlap. They all share the same style code to be style consistent. 
To get a higher resolution and show more details within the existing GPU resource constraint, we also train our model using images of $256 \times 256$ and 4 frames per batch. 

The $\lambda$ parameters. Video interpolation loss weight $\lambda_{interp}$ is set to 10. Cycle consistency loss weight $\lambda_{cycle}$ is set to 10. Style reconstruction loss weight $\lambda_{rec}$ is set to 0.025.

\subsection{Human Preference Score}

We have conducted human subjective experiments to evaluate the visual quality of synthesized videos using the Amazon Mechanical Turk (AMT) platform. 

For the video-level evaluation, we show two videos (synthesized by two different models) to AMT participants, and ask them to select which one looks more realistic regarding a video-consistency and video quality criteria. 

\begin{itemize}
    \item \textbf{UVIT (ours) / 3DCycleGAN:} Since 3DcycleGAN~\cite{bashkirova2018unsupervised} generates consistent output with 8 frames in the original paper setting, UVIT results are organized to 8 frames for a fair comparison. 
    \item \textbf{UVIT (ours) / Improved ReCycleGAN:} When comparing with improved RecycleGAN~\cite{bansal2018recycle}, we take each video clip with 30 frames. 
    \item \textbf{UVIT (ours) / vid2vid:} When comparing with vid2vid~\cite{wang2018video}, we take each video clip with 28 frames, following the setting in vid2vid~\cite{wang2018video}. 
\end{itemize}

For the image-level evaluation, we show to AMT participants two generated frames synthesized by two different algorithms, and ask them which one looks more real in visual quality. 

These evaluations have been conducted for 100 videos and frame samples to assess the image-level and video-level qualities, respectively. We gathered answers from 10 different workers for each sample.

\section{Additional examples of the  label-to-image  qualitative  comparison ($128 \times 128$)}
\begin{figure*}[htbp]
\centering   
\includegraphics[scale=0.5]{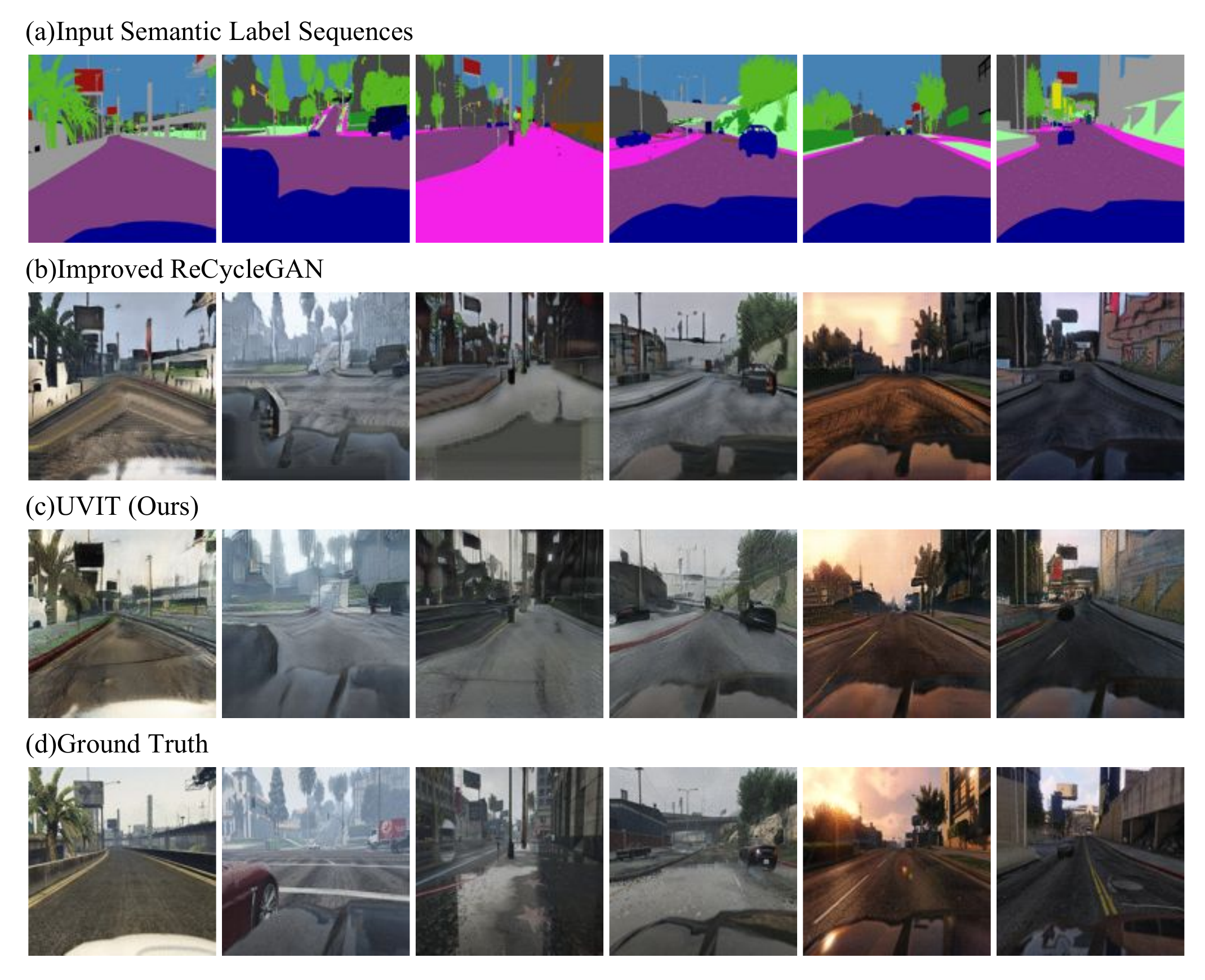}  
\caption{\textbf{Video screen cut of the label-to-image qualitative comparison. } First row: semantic label inputs; Second row: improved ReCycleGAN outputs; Third row: UVIT outputs. Fourth row: ground truth. A full video file can be found in \textit{1\_LRCompare.mp4}.}
\label{fig:ours_comparision_append}
\end{figure*}

More results on the label-to-image mapping comparison of UVIT (ours) and Improved ReCycleGAN are depicted in  \fref{fig:ours_comparision_append} and the attached video \textit{Compare.mp4}. 
For the \textit{1\_LRCompare.mp4}, we give a short description to guide the comparison. From left to right, there are outputs for six different input samples to compare:

\begin{itemize}
    \item \textbf{1:} Please see the trajectory of the car and the surrounding road.
    \item \textbf{2:} Please see the boundary between two cars.
    \item \textbf{3:} Please see the translation of the road to check the complete translation and consistency across frames.
    \item \textbf{4:} Please see the walls and the pillar across frames. 
    \item \textbf{5:} Please see the consistency of the road. 
    \item \textbf{6:} Please see the consistency of the wall.
\end{itemize}

\section{Higher resolution results ($256 \times 256$)}

To get a higher resolution and show more details within the existing GPU resource constraint, we also train our model using images of $256 \times 256$ and 4 frames per batch. During the test time, we divide a longer sequence into sub-sequences of 4 frames with overlap. All the results of this section are trained using images of $256 \times 256$ and 4 frames per batch.

\subsection{Additional examples of the label-to-image  qualitative  comparison} 
\begin{figure}
    \centering
    \includegraphics[width=\linewidth,trim={8 300 100 60},clip]{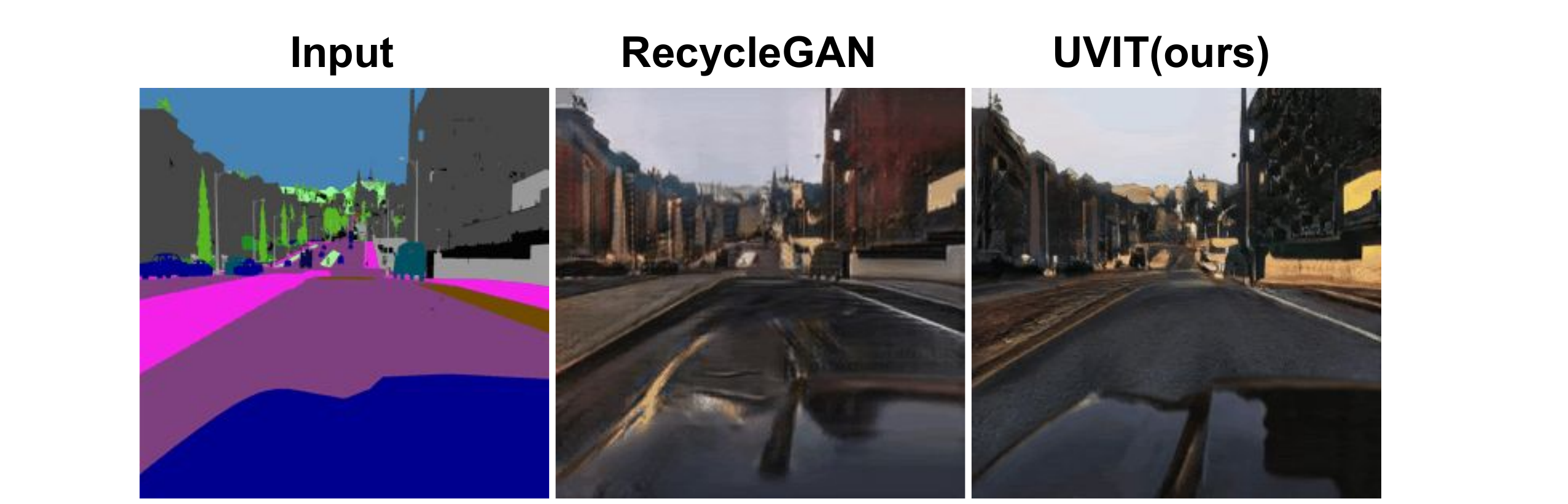}
    \caption{ \textbf{Video screenshot of the video corresponding to Fig. 1 in the main paper.} The video is attached as $2\_HRcomapare.mp4$}
    \label{fig:FIGURE1}
\end{figure}

In \fref{fig:FIGURE1} (corresponding video $2\_HRcompare.mp4$) and \fref{fig:FIGURE1} (corresponding video $3\_HRcomapare2.mp4$), we provide the visual examples of how our UVIT method compares with respect to RecycleGAN~\cite{bansal2018recycle}. The RecycleGAN outputs are generated by the original code provided by the author of RecycleGAN in $256 \times 256$. Besides the video-level quality comparison from videos, we encourage the reader to also check the frame-level quality from images since \textit{.mp4} format may fail to preserve some image-level quality.

\begin{figure}
    \centering
    \includegraphics[width=0.9\linewidth,trim={8 230 40 0},clip]{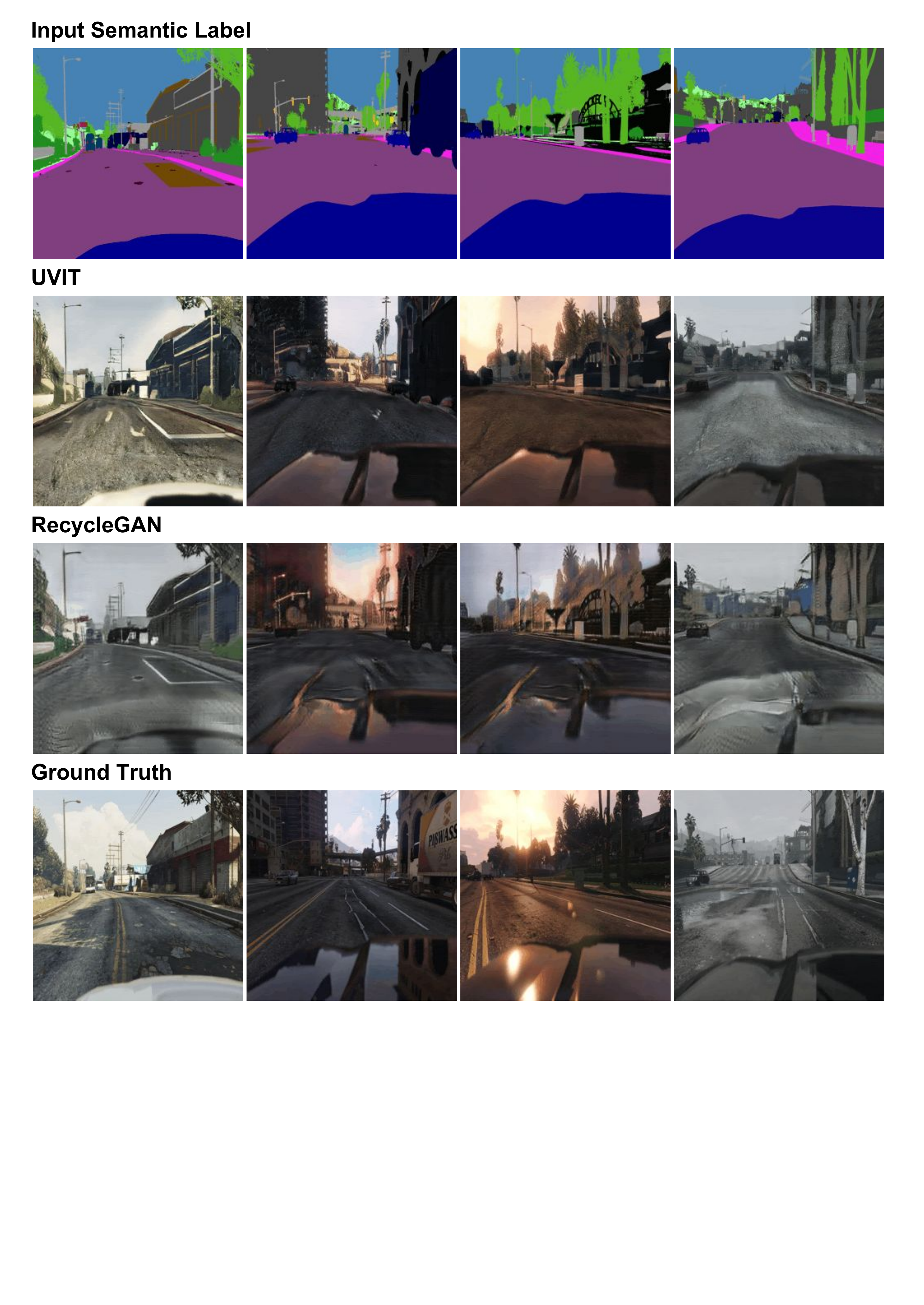}
    \caption{\textbf{Video screenshot of the comparison with RecycleGAN~\cite{bansal2018recycle}:} We aim to compare the content consistency and image-level quality. Here the RecycleGAN results are produced by the original RecycleGAN code in a resolution of $256 \times 256$. Since there is no guarantee of style consistency for RecycleGAN, we select some RecycleGAN visual results  in a small sequence length of 30 frames where style is almost consistency to compare with UVIT (ours). The corresponding video is attached as $3\_HRcomapare2.mp4$}
    \label{fig:FIGURE2}
\end{figure}

\subsection{Quantitative  comparison of the label-to-image and image-to-label}

In \tref{table:comparison_append} and \ref{table:comparison_FID_append}, we show quantitative results for our proposed method trained with a resolution of $256 \times 256$ and 4 frames per batch.

\begin{table*}[t]
\centering
  \caption{\textbf{Quantitative comparison between \SHORTTITLE{} and baseline approaches on the image-to-label (Semantic segmentation) task.($256 \times 256$ with 4 frames per batch during training)}  .Our translator effectively leverage the temporal information directly, thus producing more semantic persevering translation outcomes} 
 \begin{tabular}{c c c c c c c c}\toprule
  Criterion&Model&Day&Sunset&Rain&Snow&Night&All\\\midrule
  \multirow{2}*{mIoU$\uparrow$} &ReCycleGAN (Reproduced) &10.32&	11.19&	11.25&	9.83&	7.73&	10.12\\
 ~& \SHORTTITLE{} (Ours) (frame 4)&\textbf{12.05} &\textbf{12.23} & \textbf{13.37}&\textbf{11.54}&\textbf{10.49}	&\textbf{11.93} \\
  \midrule 
  
   \multirow{2}*{AC$\uparrow$}&ReCycleGAN (Reproduced) &15.80&	15.79&	15.93&	15.57&	11.47&	14.85\\
   ~& \SHORTTITLE{} (Ours) (frame 4)&\textbf{17.21} &\textbf{17.41} & \textbf{18.16}& \textbf{17.37}	&\textbf{14.30}& \textbf{16.50}\\
  \midrule
  
    \multirow{2}*{PA$\uparrow$} &ReCycleGAN (Reproduced) &54.70&	55.92&	57.71&	50.85&	49.11&	53.66\\
     ~& \SHORTTITLE{} (Ours) (frame 4)&  \textbf{63.44}& \textbf{61.98}& \textbf{64.72}&	\textbf{60.83}&	\textbf{62.05}& \textbf{62.35}\\
  
  \bottomrule
 \end{tabular}
 \label{table:comparison_append}
    \vspace{-0.3cm}
\end{table*}

\begin{table}[t]
\centering
  \caption{\textbf{Quantitative comparison between \SHORTTITLE{} and baseline approaches on the label-to-image task ($256 \times 256$ with 4 frames per batch during training)}. Better FID indicates that our translation has better visual quality and temporal consistency. We use the pre-trained network (I3D~\cite{carreira2017quo}) to extract features from 30-frame sequences just as the experiments in the main paper. }
 \begin{tabular}{c c c c c c c}\toprule
  Criterion&Model&Day&Sunset&Rain&Snow&Night\\\midrule
  \multirow{2}*{FID$\downarrow$}&ReCycleGAN~\cite{bansal2018recycle} &23.60 &24.45 &28.54 &31.58 & 35.74\\
  ~&\SHORTTITLE{} (ours) (frame 4) &\textbf{18.68}&\textbf{16.70}&\textbf{20.20}&\textbf{18.27}& \textbf{19.29} \\
  \bottomrule
 \end{tabular}
   \label{table:comparison_FID_append}
\end{table}

\subsection{Label-to-image multi-subdomain and multimodality results}
\begin{figure}[ht]
\centering
\includegraphics[width=0.8\linewidth]{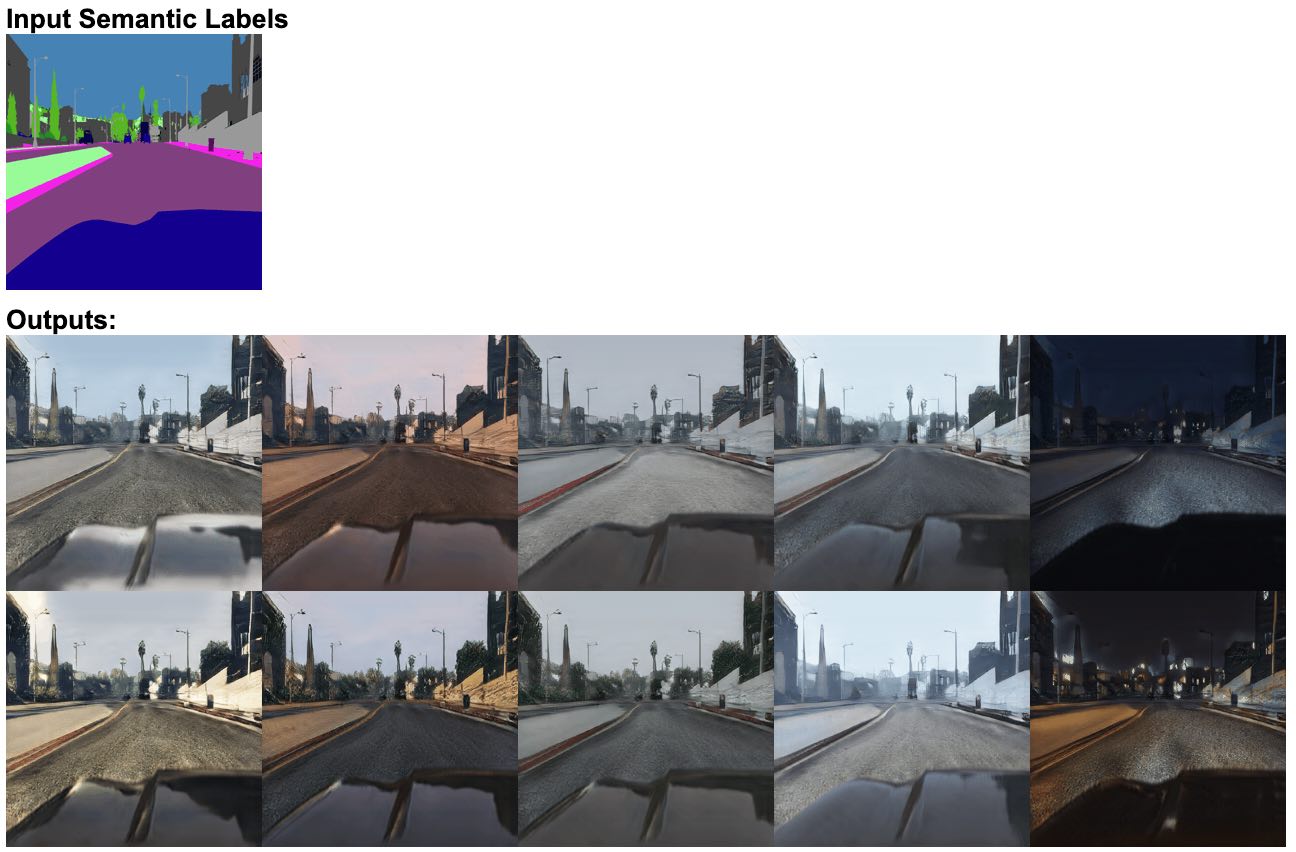}
  \caption{\textbf{Video screenshot of the label-to-image multi-subdomain and multimodality results}. Better depicted in \textit{4\_Multimodality.mp4}.}
\label{fig:Multimodaliy_append}
\end{figure}

Video results of UVIT on label sequences to image sequences with multi-subdomain and multimodality are shown in \fref{fig:Multimodaliy_append} and the enclosed video \textit{4\_Multimodality.mp4}. The videos are all with a length of 220 frames.

\subsection{Long video example (1680 frames)}

\begin{figure}
    \centering
    \includegraphics[width = 0.25 \linewidth]{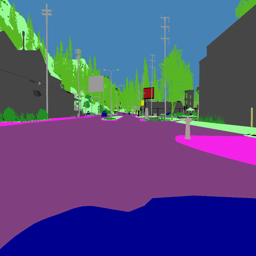}
    \includegraphics[width = 0.25 \linewidth]{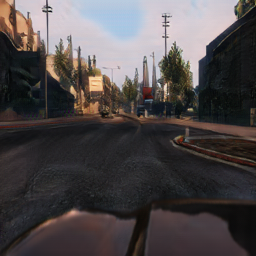}
    \caption{\textbf{Screenshot of a long style consistent translation video visual example (1680 frames)}. Left: input semantic labels; Right: UVIT translated video in sunset scenario. All frames within the video share the same style code to keep style consistency. The video is attached as $5\_long\_consistency.mp4$ }
    \label{fig:long}
\end{figure}

In \fref{fig:long} and attached video $5\_long\_consistency.mp4$, we provide a video sequence example with more than 1680 frames to give a qualitative example of how our UVIT model performs in terms of style consistency. Note that the semantic labels in Viper~\cite{richter2017playing} are automatic generated, however, we observe that there may still exist a little small flips in the input semantic label sequence occasionally. 

\clearpage
\subsection{Translation on other datasets}

\begin{figure}
    \centering
    \vspace{0.5cm}
    \includegraphics[width=0.9\linewidth,trim={0 260 10 0},clip]{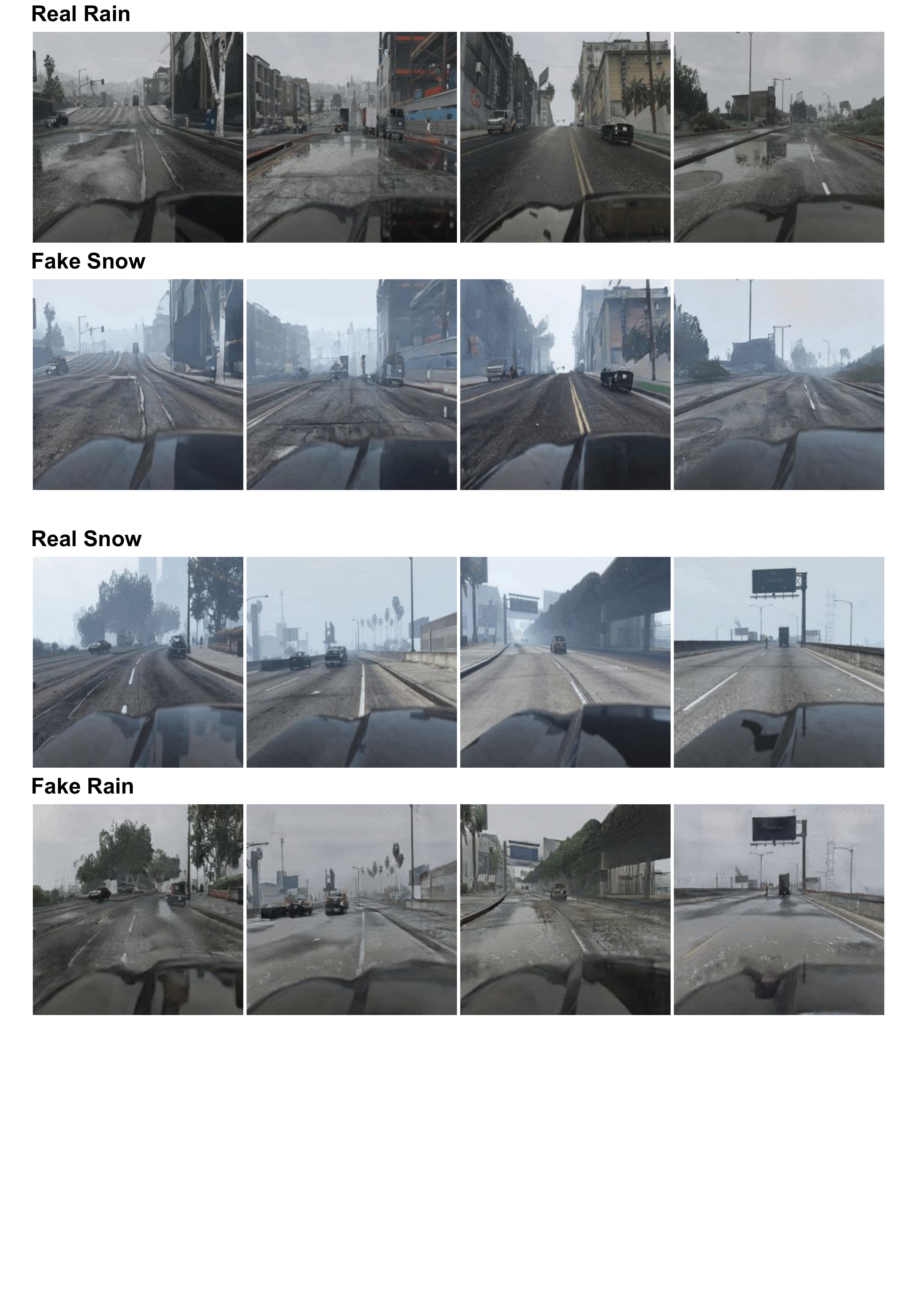}
    \caption{\textbf{Screenshot of Viper Rain-and-Snow translation.} First row: real rain inputs; Second row: translated snow videos; Third row: real snow inputs; Fourth row: translated rain videos. Video is attached as \textit{6\_Rainandsnow.mp4}}
    \label{fig:R2S_HR_append}
\end{figure}

\begin{figure}
    \centering
    \includegraphics[width=0.9\linewidth,trim={0 260 10 0},clip]{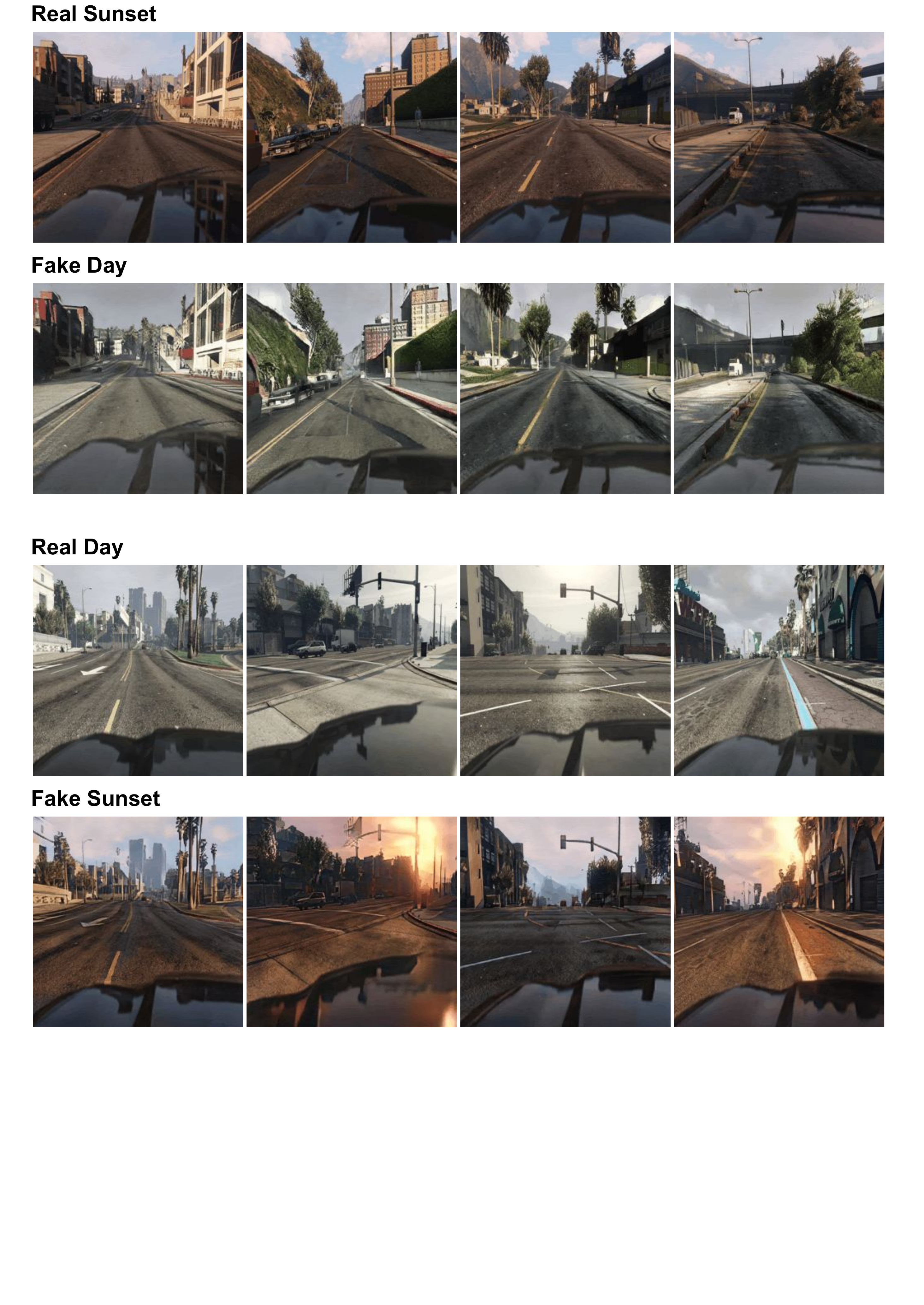}
    \caption{\textbf{Screenshot of Viper Sunset-and-Day translation.}  First row: real sunset inputs; Second row: translated day videos; Third row: real day inputs; Fourth row: translated sunset videos. Video is attached as \textit{7\_Sunsetandday.mp4}}
    \label{fig:S2D_HR_append}
\end{figure}

\begin{figure}
    \centering
    \includegraphics[width=0.9\linewidth]{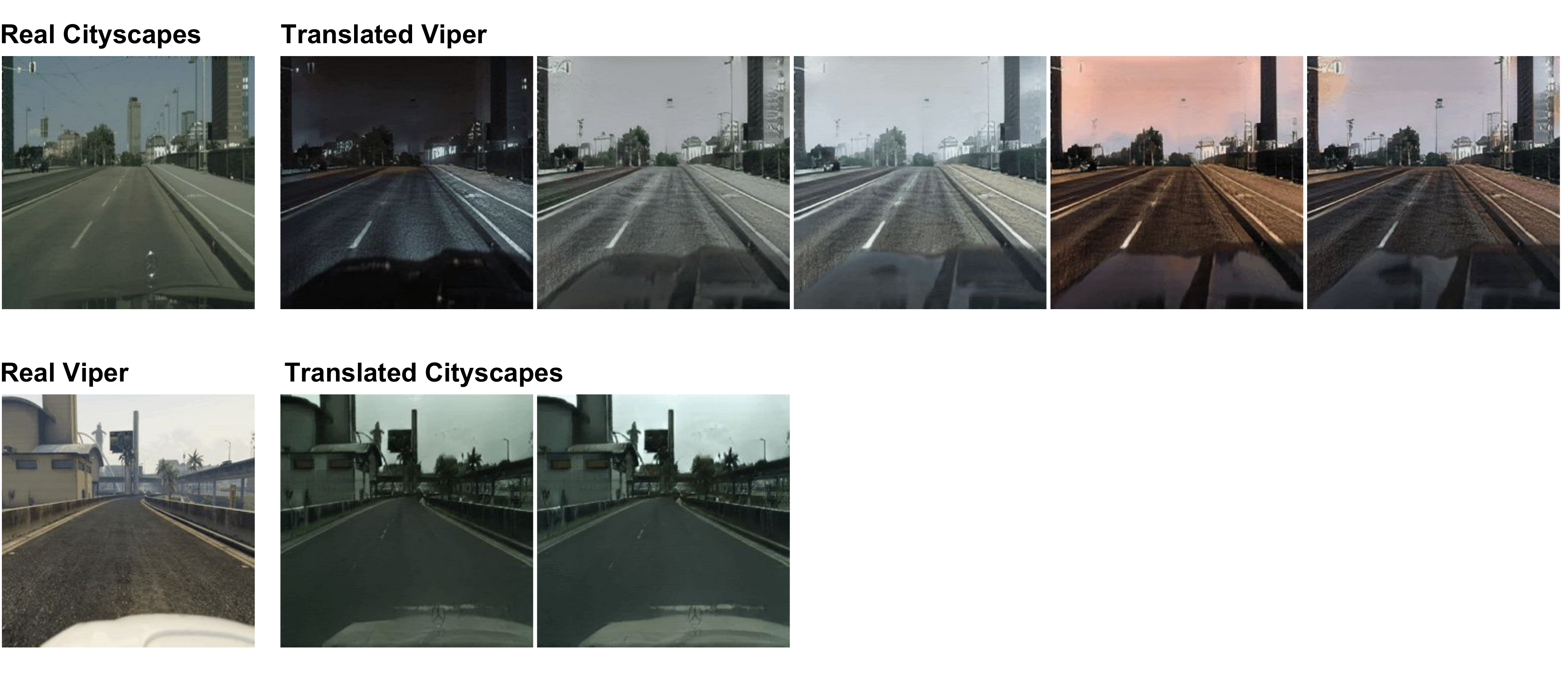}
    \caption{\textbf{Screenshot of Cityscapes-and-Viper translation.}  Top left: real Cityscapes input; Top right: translated Viper videos with different style codes; Bottom left: real Viper input; Bottom right:  translated Cityscapes videos with different style codes.  Since the general distribution between Cityscapes and Viper may be different (\emph{e.g. } there are more buildings in Cityscapes), the translated Viper video may differ from input Cityscapes video in class distribution to fool the discriminator, so as to be close to the class distribution in the target domain. Video is attached as \textit{8\_Cityscapesandviper.mp4}}
    \label{fig:C2G_HR_append}
\end{figure}

In \fref{fig:R2S_HR_append} and in the attached video \textit{6\_Rainandsnow.mp4}, we provide visual examples of UVIT video translation between Rain and Snow scenes in the Viper dataset. In \fref{fig:S2D_HR_append} and in the attached video \textit{7\_Sunsetanday.mp4}, we provide visual examples of UVIT video translation between Sunset and Day scenes in the Viper dataset. In \fref{fig:C2G_HR_append} and in the attached video \textit{8\_Cityscapesandviper.mp4}, we provide visual examples of UVIT video translation between Cityscapes dataset and Viper dataset. Besides the video-level quality evaluation from videos, we encourage the reader to also check the frame-level quality from images since \textit{.mp4} format may fail to preserve some image-level quality.

\clearpage
\section{Additional Results for Ablation Study}
\label{sec:experimentsablationappend}

Here we provide the supplementary results for the ablation part. We first give a complete table for the ablation experiment on how the performance depends on the input frame number. After that, we give the qualitative results of multimodal consistent videos when UVIT is trained and tested without the sub-domain label.

\subsection{Different input frame number} Our RNN-based translator incorporates temporal information from multiple frames. In \tref{Ablation2_append}, we provide the complete table corresponding to the "Different input frame number" ablation study section in the main paper.

\subsection{UVIT without the subdomain label}

\begin{figure}
    \centering
    \includegraphics[width=0.9\linewidth,clip]{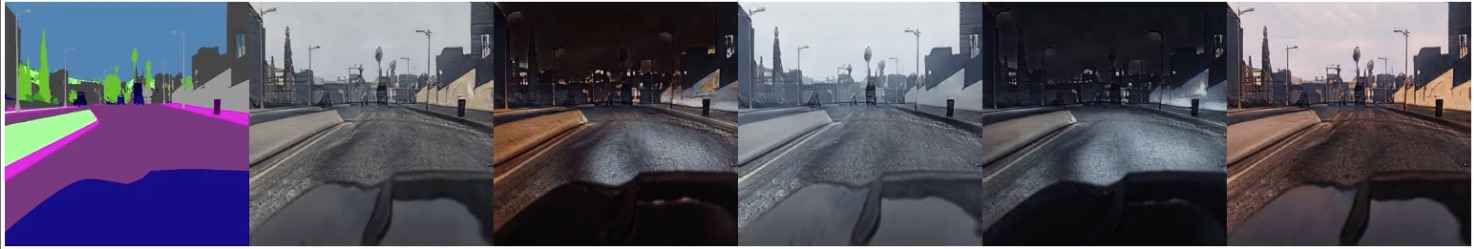}
    \caption{\textbf{Ablation study: when no sub-domain label is used during training and testing.} First video is the input semantic label sequence, the rest videos are the translated scene videos with style codes randomly sampled from prior distribution. There are 220 frames for each video. The corresponding video is attached as ${9\_No\_subdomain.mp4}$}
    \label{fig:No_sub}
\end{figure}

To check how our UVIT performs in terms of style consistency without the subdomain label information, we run this ablation experiment. The results are attached in \fref{fig:No_sub} and ${9\_No\_subdomain.mp4}$. By randomly sampling the style code from prior distribution, we can get multimodal consistent video results in a stochastic way.

\begin{table*}[t]
\centering
  \caption{\textbf{Quantitative results of  \SHORTTITLE{} with different number of frames per batch in training on the image-to-label (Semantic segmentation) task.} With the increase of input frames number, our RNN-based translator can utilize the temporal information better, resulting in better semantic preserving}
 \begin{tabular}{c c c c c c c c}\toprule
  Criterion&Frame number&Day&Sunset&Rain&Snow&Night&All\\\midrule
  \multirow{4}*{mIoU$\uparrow$}&4 &11.84&	11.91&	12.35&	11.37&	8.49&	11.19\\
  ~&6 &{12.29}&12.66&{13.03}&{11.77}&	{9.79}&{11.94}\\
  ~& 8 &{13.05}&13.21&14.23&{13.07}&	\textbf{11.00} &{12.87}\\
  ~& 10 &\textbf{13.71} &\textbf{13.89} & \textbf{14.34}&\textbf{13.23}&{10.10}	&\textbf{13.07} \\
  \midrule
  
   \multirow{4}*{AC$\uparrow$}& 4 &16.78&	16.75&	16.57&	16.32&	12.21&	15.7\\
   ~& 6 &{17.50}&{17.46}&	{17.66}&	{16.73}& {14.23}&	{16.62}\\
~& 8 &{18.42}&{18.28}&	\textbf{19.19}&	{17.80}&	\textbf{15.18}&\textbf{17.68}\\
 ~& 10 &\textbf{18.74} &\textbf{19.13} & {18.98}& \textbf{17.81}	&{13.99}& {17.59}\\
  \midrule
  
    \multirow{4}*{PA$\uparrow$}& 4 &62.84&	60.34&	61.97&	58.77&	51.68&	59.04\\
    &6 &{62.85}&	61.21&	{62.21}&	{59.77}&	56.84&{60.51}\\
  ~& 8 &{65.56}&{64.11}&{66.26}&{64.18}&	\textbf{ 62.02}& {64.25}\\
 ~& 10 &  \textbf{68.06}&  \textbf{66.35}& \textbf{67.21}&	\textbf{65.49}&	{58.97}& \textbf{65.20}\\
  
  \bottomrule
 \end{tabular}
  \label{Ablation2_append}

\end{table*}

\end{document}